\def\paperID{5686}
\def\confName{CVPR}
\def\confYear{2024}

\def\paperTitle{GALA: Generating Animatable Layered Assets from a Single Scan}

\def\authorBlock{
    Taeksoo Kim\textsuperscript{1}\thanks{Equal contribution} \qquad
    Byungjun Kim\textsuperscript{1}\footnotemark[1] \qquad
    Shunsuke Saito\textsuperscript{2} \qquad
    Hanbyul Joo\textsuperscript{1}
    \and
    \textsuperscript{1}Seoul National University \qquad
    \textsuperscript{2}Codec Avatars Lab, Meta
    \and
    {\tt\small \{taeksu98, byungun.kim, hbjoo\}@snu.ac.kr \quad shunsukesaito@meta.com} \\
    {\tt\small \href{https://snuvclab.github.io/gala/}{\color{magenta}{https://snuvclab.github.io/gala/}}}
}

\newif\ifreview 
\newif\ifarxiv \newcommand{\arxiv}{\arxivtrue}
\newif\ifcamera 
\newif\ifrebuttal 

\arxiv %

\pdfoutput=1
\documentclass[10pt,twocolumn,letterpaper]{article}
\ifreview \usepackage[review]{cvpr} \fi
\ifarxiv \usepackage[pagenumbers]{cvpr} \fi
\ifrebuttal \usepackage[rebuttal]{cvpr} \fi
\ifcamera \usepackage{cvpr} \fi

\usepackage{graphicx}
\usepackage{amsmath}
\usepackage{amssymb}
\usepackage{booktabs}

\usepackage{times}
\usepackage{microtype}
\usepackage{epsfig}
\usepackage[table,xcdraw]{xcolor}
\usepackage{caption}
\usepackage{float}
\usepackage{placeins}
\usepackage{color, colortbl}
\usepackage{stfloats}
\usepackage{enumitem}
\usepackage{tabularx}
\usepackage{xstring}
\usepackage{multirow}
\usepackage{xspace}
\usepackage{url}
\usepackage{subcaption}
\usepackage{xcolor}
\usepackage{cuted}
\usepackage[hang,flushmargin]{footmisc}
\usepackage{bm}

\usepackage{natbib}
\usepackage[normalem]{ulem}
\usepackage{breqn}
\usepackage{pifont}%
\newcommand{\cmark}{\ding{51}}%
\newcommand{\xmark}{\ding{55}}%
\renewcommand{\vec}[1]{\bm{#1}}
\newcommand{\mat}[1]{\mathbf{#1}}
\newcommand{\mesh}[1]{\mathcal{#1}}

\definecolor{lavender}{HTML}{F49EC4}
\definecolor{apricot}{HTML}{FBB982}
\newcommand*{\boldone}{\text{\usefont{U}{bbold}{m}{n}1}}
\usepackage{indentfirst}
\usepackage{multibib}
\usepackage{xpatch}

\ifcamera \usepackage[accsupp]{axessibility} \fi

\ifarxiv  \fi

\DeclareMathOperator*{\argmin}{arg\,min}

\newcommand{\R}[1]{{%
    \textbf{%
        \ifstrequal{#1}{1}{\textcolor{red}{R#1}}{%
        \ifstrequal{#1}{2}{\textcolor{blue}{R#1}}{%
        \ifstrequal{#1}{3}{\textcolor{magenta}{R#1}}{%
        \ifstrequal{#1}{4}{\textcolor{teal}{R#1}}{%
                           \textcolor{cyan}{R#1}%
        }}}}%
    }%
}}

\newcites{main}{References} %
\newcites{supp}{References} %

\usepackage{xr-hyper}

\makeatletter
\newcommand*{\addFileDependency}[1]{
  \typeout{(#1)}
  \@addtofilelist{#1}
  \IfFileExists{#1}{}{\typeout{No file #1.}}
}

\makeatother

\usepackage[pagebackref,breaklinks,colorlinks]{hyperref}
\usepackage[capitalize]{cleveref}
\crefname{section}{Sec.}{Secs.}
\crefname{table}{Table}{Tables}
\crefname{figure}{Fig.}{Figs.}

\begin{document}
\title{\paperTitle}
\author{\authorBlock}
\maketitle

\begin{strip}
\vspace*{-2.0cm}
\centering
\includegraphics[width=1\textwidth, trim={1cm 0 0 0}]{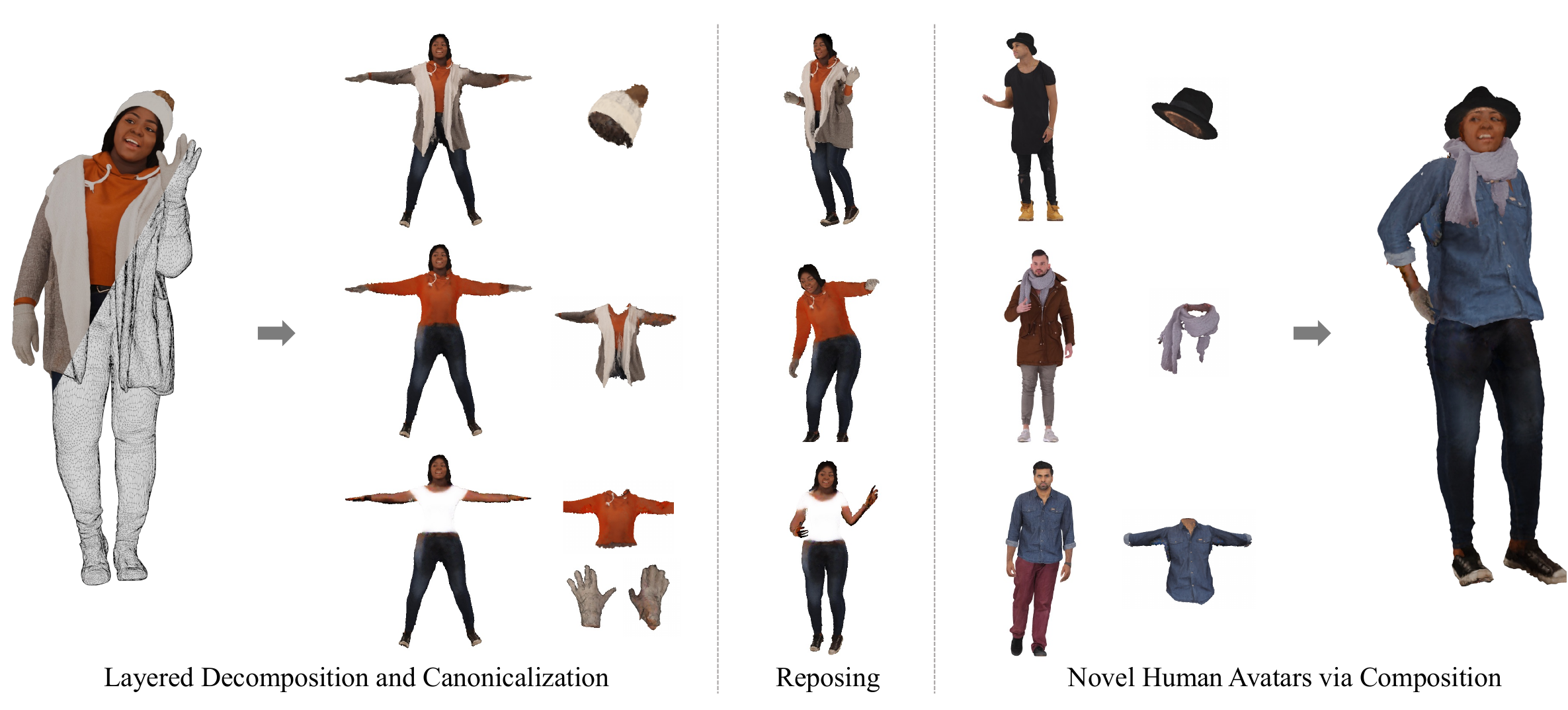}
\vspace{-20pt}
\captionof{figure}{\textbf{GALA.} Given a single-layer 3D mesh of a clothed human (left), our approach enables \textbf{G}eneration of \textbf{A}nimatable \textbf{L}ayered \textbf{A}ssets for 3D garment transfer and avatar customization in any poses by decomposing and inpainting the geometry and texture of each layer with a pretrained 2D diffusion model in a canonical space.
}
\label{fig:teaser}
\end{strip}%

\begin{abstract}

We present \textbf{GALA}, a framework that takes as input a single-layer clothed 3D human mesh and decomposes it into complete multi-layered 3D assets. The outputs can then be combined with other assets to create novel clothed human avatars with any pose.
Existing reconstruction approaches often treat clothed humans as a single-layer of geometry and overlook the inherent compositionality of humans with hairstyles, clothing, and accessories, thereby limiting the utility of the meshes for down-stream applications.
Decomposing a single-layer mesh into separate layers is a challenging task because it requires the synthesis of plausible geometry and texture for the severely occluded regions. Moreover, even with successful decomposition, meshes are not normalized in terms of poses and body shapes, failing coherent composition with novel identities and poses. 
To address these challenges, we propose to leverage the general knowledge of a pretrained 2D diffusion model as geometry and appearance prior for humans and other assets. We first separate the input mesh using the 3D surface segmentation extracted from multi-view 2D segmentations.
Then we synthesize the missing geometry of different layers in both posed and canonical spaces using a novel pose-guided Score Distillation Sampling (SDS) loss.
Once we complete inpainting high-fidelity 3D geometry, we also apply the same SDS loss to its texture to obtain the complete appearance including the initially occluded regions.
Through a series of decomposition steps, we obtain multiple layers of 3D assets in a shared canonical space normalized in terms of poses and human shapes, hence supporting effortless composition to novel identities and reanimation with novel poses.  
Our experiments demonstrate the effectiveness of our approach for decomposition, canonicalization, and composition tasks compared to existing solutions.

\end{abstract}

\section{Introduction}
\label{sec:intro}

In the era where social interactions become increasingly online, the ability to customize digital representations of oneself is more important than ever. This is particularly critical in the domain of virtual try-on and photorealistic avatar customization. However, creating assets that can be easily layered on top of any avatars typically requires substantial manual efforts by artists. Our goal is to enable automatic creation of reusable 3D layered assets that can be effortlessly composed to any human with any poses. 

Unlike artist-created 3D assets, reconstruction-based 3D models are getting widely accessible.  
In addition to online market places of high-quality 3D scans~\cite{renderpeople, axyz}, single-view reconstruction methods~\cite{saito2019pifu,saito2020pifuhd, alldieck2022phorhum} or text-to-3d generation techniques~\cite{poole2023dreamfusion, lin2023magic3d, chen2023fantasia3d} further simplify the creation of 3D models. 
Despite these advancements, using these 3D models for virtual try-on or avatar customization remains an open challenge because these models are typically single-layer and not animatable. Different attributes such as hair, clothing, and accessories are glued into a single triangle mesh, and anything beneath the outermost layer is fully occluded. Moreover, self-contact regions are also connected, making re-animation challenging.

To address this, we propose a fully automatic framework for creating compositional layered 3D assets from a single-layer scan. Unlike the existing text-based 3D generation methods~\cite{poole2023dreamfusion,lin2023magic3d} that only support the generation of each asset in isolation, our approach learns to decompose a mesh into multiple layers and inpaint missing geometry and appearance for compositing the decomposed assets into novel identities. 
Our key idea is to complement missing geometric and appearance information by leveraging a strong image prior built from a large-scale image collections.
In particular, we leverage a latent diffusion model~\cite{rombach2022latent} that is trained on an extremely large corpus of images.
Using a score distillation sampling (SDS), we inpaint the occluded regions while retaining the originally visible regions.

For reposing, simply inpainting the geometry and appearance in an input posed space is not sufficient. For garment transfer across different identities with various poses, we need to represent the target asset and the remaining human layer in individual canonical spaces. 
However, we observe that the vanilla SDS loss often provides poor guidance by ignoring the target pose information.
We address the lack of pose-sensitivity in the SDS loss by introducing a pose-guided SDS loss. 
Specifically, we derive the SDS loss with a pose-conditioned diffusion model~\cite{zhang2023controlnet}.
This allows us to supervise the shape and appearance jointly in both posed and canonical spaces.
Once we obtain the canonicalized object and human layers, we can mix and match with other assets to create virtual try-on as shown in \cref{fig:teaser}. The composite results are further refined with penetration handling.

As there is no established benchmark for decomposition, canonicalization, and composition from a single scan, we establish a new evaluation protocol to quantitatively assess our approach. For decomposition, our approach significantly outperforms recent text-driven 3D editing methods. We also show that the proposed pose-guided SDS enables robust canonicalization even for challenging cases, outperforming existing methods. Lastly, we show garment transfer to create novel avatars only from a collection of single-layer clothed humans. Our contributions can be summarized as follows:
\begin{itemize}[itemsep=0pt]
    \item We propose a new task of multi-layer decomposition and composition from a single-layer scan, which offers a practical compositional asset creation pipeline.
    \item We present a pose-guided SDS loss, enabling the robust modeling of layered clothed humans in a canonical space for garment transfer and reposing from a single scan.
    \item We provide a comprehensive analysis of generating animatable layered assets from a single scan with a newly established evaluation protocol. We will release code for benchmarking future research on this novel task.
\end{itemize}

\section{Related Work}
\label{sec:related}

\noindent \textbf{Clothed Human Modeling.}
3D parametric human models~\cite{loper2015smpl, pavlakos2019smplx, joo2018total, xu2020ghum} have been proposed to model diverse poses and shapes of humans, allowing us to reconstruct minimally clothed 3D humans~\cite{bogo2016keep,kanazawa2017hmr, pavlakos2019smplx, rong2021frankmocap, zhang2021pymaf}.
To represent clothed humans, follow-up work leverages 3D displacements on top of the template body model~\cite{ma2020cape, alldieck2019learning, alldieck2018video}, or separate mesh layers~\cite{bhatnagar2019multi, pons2017clothcap}. Yet, the topological constraints and the resolution of the template model limit their ability to model clothing with complex shapes and high-frequency details.
In recent years, deep implicit shape representations~\cite{chen2019learning, mescheder2019occupancy, park2019deepsdf, mildenhall2020nerf, xie2022neural} have emerged as a significant breakthrough in modeling 3D humans, demonstrating their efficacy in reconstructing detailed clothed humans from images, scans, depth maps, or pointclouds~\cite{saito2019pifu, saito2020pifuhd, zheng2021pamir, tao2021function4d, xiu2022icon, dong2022pina, tiwari2021neural, saito2021scanimate, MetaAvatar:NeurIPS:2021, mihajlovic2022coap}.
Extending work enables the animations of these reconstructions~\cite{saito2021scanimate, chen2021snarf, chen2022gdna, tiwari2021neural, MetaAvatar:NeurIPS:2021, mihajlovic2021leap, mihajlovic2022coap, dong2022pina} by learning a canonical 3D shape in a space normalized in terms of human poses and shapes.
Since these approaches treat the clothed human as a single-layer mesh, several work~\cite{bertiche2020cloth3d, patel20tailornet, vidaurre2020fully, chen2021tightcap, pons2017clothcap} attempts to model the clothing of humans as a separate layer. SMPLicit~\cite{corona2021smplicit} models clothing with implicit shape representation on top of the parametric mesh model. ReEF~\cite{zhu2022reef} registrates template meshes to implicit surfaces.
There are a few attempts to enable compositional and animatable modeling of avatars.
SCARF~\cite{Feng2022scarf} separately models humans and clothing from video observations using a hybrid representation of mesh and NeRF~\cite{mildenhall2020nerf}. MEGANE~\cite{li2023megane} models high-fidelity compositional heads and eyeglasses from multi-view videos. NCHO~\cite{kim2023ncho} learns compositional generative models of humans and objects from multiple scans with and without objects in an unsupervised manner. 
Unlike existing approaches, our approach enables the modeling of animatable \emph{multi-layer} assets from a \emph{single scan}. To enable this, we exploit an image prior from a pretrained diffusion model~\cite{rombach2022latent}.

\noindent \textbf{3D Content Generation.}
Recent advancements in 3D representations~\cite{mildenhall2020nerf,xie2022neural} and generative modeling~\cite{nips2014gan,ho2020ddpm} have spurred active research for 3D content generation. Generation from text, in particular, has gained popularity due to its intuitive interface.
Early work like Text2Shape~\cite{chen2018text2shape} trains text and shape encoders to learn joint embeddings, generating text-consistent 3D shapes.
Due to the challenges of collecting large-scale paired text-3D datasets, several approaches~\cite{khalid2022clipmesh, michel2022text2mesh, jain2022dreamfields, hong2022avatarclip} utilize pretrained CLIP model~\cite{radford2021clip} for text-guided 3D content generation. 
With the recent rise of diffusion models~\cite{ho2020ddpm, song2021scorebased} for high-quality image generation~\cite{dhariwal2021diffusion,rombach2022latent}, DreamFusion~\cite{poole2023dreamfusion} proposes score distillation sampling (SDS) loss for optimizing 3D scenes represented as NeRF~\cite{mildenhall2020nerf} by leveraging the 2D diffusion prior. Various 3D representations such as point clouds~\cite{nichol2022pointe, zeng2022lion}, meshes~\cite{Liu2023MeshDiffusion, lin2023magic3d, chen2023fantasia3d}, and neural fields~\cite{metzer2023latentnerf, seo2023let} have also been utilized for 3D generation.
Some approaches~\cite{kim2023chupa, huang2023humannorm, wang2023rodin, hu2023humanliff, shi2023MVDream} incorporates additional 3D datasets with diffusion model to enable high-quality 3D generation. MVdream~\cite{shi2023MVDream} generates multi-view images by finetuning the diffusion model with multi-view rendering of Objaverse~\cite{deitke2023objaverse}. Chupa~\cite{kim2023chupa} and HumanNorm~\cite{huang2023humannorm} finetune the diffusion model to generate normal or depth maps for generating 3D humans with fine geometric details.
However, current 3D content generation methods generate 3D assets \emph{as a single-layer mesh}, limiting their utility for composition with other assets. In contrast, our approach leverages the 2D diffusion prior to create decomposed layers of attributes in a canonical space, facilitating garment transfer and reposing.

\noindent \textbf{3D Editing.}
Editing 3D scenes has traditionally been a task for experienced artists, but recent work shows the great potential of text-based automatic 3D content manipulation.
Instruct-NeRF2NeRF~\cite{hqaue2023in2n} edits the pretrained NeRF using prompt by iteratively updating training images of the NeRF through Instruct-Pix2Pix~\cite{brooks2022instructpix2pix}. DreamEditor~\cite{zhuang2023dreameditor} exploits mesh-based neural fields~\cite{yang2022neumesh} to enable local and flexible editing via SDS loss~\cite{poole2023dreamfusion} using a diffusion model finetuned with DreamBooth~\cite{ruiz2023dreambooth}. 
Vox-E~\cite{sella2023voxe} similarly utilizes SDS loss but enables local editings using the 3D attention map aggregated from the 2D attention maps of a diffusion model. FocalDreamer~\cite{li2023focaldreamer} employs an additive approach to edit the geometry of input 3D scans, creating reusable independent assets. While our approach shares the motivation of FocalDreamer~\cite{li2023focaldreamer} in the sense of generating reusable 3D assets, our method does not require the designated editing region as an additional input and focuses on the \emph{decomposition} of the input 3D scan into multiple reusable layers instead of the addition of new components.

\begin{figure*}[t]
\includegraphics[width=0.98\linewidth, trim={0cm 0cm 1cm 0cm}]{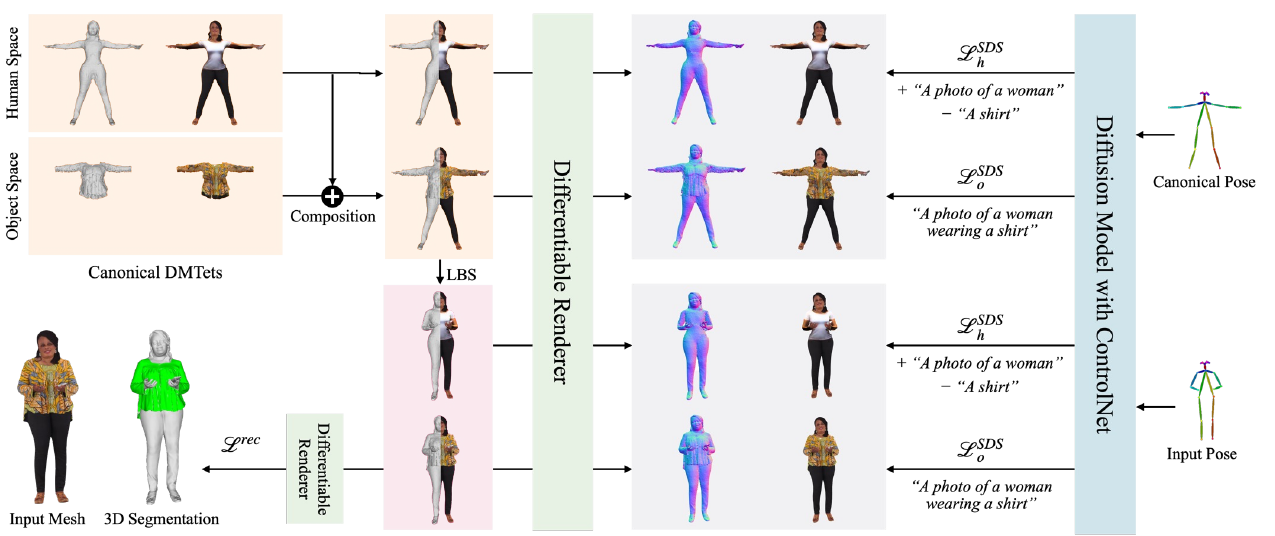}
\caption{\textbf{Overview.} GALA learns an object and the remaining human layers in a canonical space using DMTet~\cite{shen2021dmtet}. The canonical space colored \textcolor{orange}{orange} and the original posed space colored \textcolor{purple}{purple} are differentiably associated with linear blend skinning (LBS). Our novel pose-guided SDS loss (right) guides the decomposition and inpainting in both the canonical and posed space. We also retain the fidelity of visible regions via a reconstruction and segmentation loss (left-bottom).}
\label{fig:overview}
\end{figure*}

\section{Preliminaries}
\label{sec:preliminaries}
\subsection{Score Distillation Sampling}
\label{sds}
To synthesize 3D scenes without requiring large-scale 3D datasets, DreamFusion~\cite{poole2023dreamfusion} introduces Score Distillation Sampling (SDS) loss. SDS loss leverages the knowledge of a pretrained 2D diffusion model. Given the target prompt, the loss optimizes over the 3D volume parameterized with $\theta$ using the differentiable renderer $g$, such that the generated image $\vec{x} = g(\theta)$ closely resembles samples from the frozen diffusion model, $\phi$. The gradient of the loss is calculated as,
\begin{equation}
\label{eq:sds}
\begin{split}
    \nabla_{\theta} \mathcal{L}^{SDS}(\vec{x}, \phi) 
    = \mathop{\mathbb{E}} \left[ \omega(t)(\hat{\epsilon_{\phi}}(\vec{x}_t; \vec{y}, t) - \epsilon) \frac{\partial \vec{x}}{\partial \theta} \right],
\end{split}
\end{equation}
where $\vec{y}$ denotes text condition and $t$ is the noise level. $\vec{x}_t$ denotes the noised image, $\hat{\epsilon_{\phi}}(\vec{x}_t; \vec{y}, t)$ represents the noise prediction for the sampled noise $\epsilon$, and $\omega(t)$ is the weighting function defined by the scheduler of the diffusion model. 

\subsection{Deep Marching Tetrahedra}
\label{dmtet}
We adopt Deep Marching Tetrahedra~\cite{shen2021dmtet} (DMTet) as our geometric representation, which is an implicit-explicit hybrid 3D representation. It employs a deformable tetrahedral grid denoted as $(X_{T}, T)$, where $X_T$ represents the grid's 3D vertices and $T$ defines the tetrahedral structure, where each tetrahedron contains four vertices in $X_{T}$.
For each vertex $\vec{x}_i \in X_T$,
DMTet predicts the signed distance value $s(\vec{x}_i)$ from the surface and the position offset $\Delta\vec{x}_i$ of each vertex and extracts a triangular mesh from the implicit field using the differentiable Marching Tetrahedral (MT) layer.
Since the pipeline is fully differentiable, losses defined explicitly on the surface mesh can be used for optimizing the surface geometry represented by DMTet.

\section{Method}
\label{sec:method}

Our method decomposes a single-layer 3D human scan into two complete layers of the target object and the rest of the scan in separate canonical spaces. Following the previous work~\cite{chen2023fantasia3d}, we first model the geometry and subsequently model the appearance, and adopt DMTet~\cite{shen2021dmtet} as our geometric representation (\cref{subsection:geometry}). 
To reconstruct visible parts of each layer, we lift multi-view 2D segmentations of the target object onto the input 3D scan. Using forward linear blend skinning (LBS), we transform the canonical geometry of each layer to the pose of the input scan and reconstruct the visible part of each layer based on the acquired segmentation.
We further leverage a 2D diffusion prior via our pose-guided SDS loss applied in canonical space to enable canonicalization of a \emph{single} scan and complete the geometry of the occluded regions(\cref{subsection:reconstructon}).
Once we optimize the geometry of the human and the object, we model the appearance using similar SDS losses in the canonical space (\cref{subsection:appearance}).
Lastly, we refine the composition of the decomposed layers by reducing self-penetration (\cref{subsection:composition}). \cref{fig:overview} shows an overview of our pipeline.

\subsection{Representation and Initialization}
\label{subsection:geometry}
We model the geometry of the human and an object in separate canonical spaces using DMTet~\cite{shen2021dmtet}. For a given tetrahedral grid for the human $(X_{T_h}, T_h)$ and for the object $(X_{T_o}, T_o)$, we utilize MLP networks $\bm{\Psi}_h$ and $\bm{\Psi}_o$ to predict the signed distance and the deformation offset of every vertex of the grids. Using the predicted signed distance and offset, the canonical human mesh, $\mesh{M}^c_h=(\mesh{V}^c_h, \mesh{F}_h)$, and the canonical object mesh, $\mesh{M}^c_o=(\mesh{V}^c_o, \mesh{F}_o)$, can be extracted from each grid via a differentiable MT layer, where $\mesh{V}^c_h$ and $\mesh{V}^c_h$ denotes the vertices, and $\mesh{F}_h$ and $\mesh{F}_o$ denotes the faces of each mesh.
To obtain a posed mesh, we transform every vertex of the reconstructed mesh via forward linear blend skinning (LBS)~\cite{saito2021scanimate, chen2021snarf}, utilizing the skinning weights of the nearest neighbor vertex of the canonical SMPL-X mesh~\cite{pavlakos2019smplx}. Formally, a vertex $\vec{v}^c \in \mesh{V}^c_h \cup \mesh{V}^c_o$ in canonical space is transformed into a posed space with,
\begin{equation}
\label{eq:warp}
    \bar{\vec{v}}^p =(\sum_{i=1}^{n_b} w_i \cdot \mat{T}_i(\vec{\beta}, \vec{\theta}))
    \cdot \begin{bmatrix} \mat{I} & \mathcal{B}(\vec{\beta}, \vec{\theta}, \vec{\psi}) \\ 0 & 1 \end{bmatrix} \cdot \bar{\vec{v}}^c,
\end{equation}
where $\bar{\vec{v}}^p, \bar{\vec{v}}^c$ are homogeneous coordinates of $\vec{v}^p, \vec{v}^c$ respectively, $n_b$ is the number of bones, $w_i$ is the blend skinning weight of the bone $i$, and $\mat{T}_i(\vec{\beta}, \vec{\theta})\in \mathbb{R}^{4 \times 4}$ is the transformation of the bone $i$ in SMPL-X model given shape parameter $\vec{\beta}\in \mathbb{R}^{10}$ and pose parameter $\vec{\theta} \in \mathbb{R}^{55 \times 3}$. Blend shapes $\mathcal{B}(\vec{\beta}, \vec{\theta}, \vec{\psi})$ are the summation of identity blend shapes, pose blend shapes, and the expression blend shapes, where $\vec{\psi}\in \mathbb{R}^{10}$ is the expression parameter. By transforming all vertices, we get the posed human mesh, $\mesh{M}^p_h=(\mesh{V}^p_h, \mesh{F}_h)$, and the posed object mesh, $\mesh{M}^p_o=(\mesh{V}^p_o, \mesh{F}_o)$. 
For ease of notation, we use $LBS(\cdot)$ to specify the relationship between the canonical mesh and posed mesh as follows:
\begin{align}
    \mesh{M}^p_{\{h, o\}}=LBS(\mesh{M}^c_{\{h, o\}}).
\end{align}

We initialize our DMTets using SMPL-X mesh in canonical pose. We sample points $\vec{q} \in \mathbb{R}^3$ in each space, compute the signed distance $SDF(\vec{q})$ from each point to the SMPL-X mesh, and optimize the following loss functions.
\begin{align}
\label{eq:init}
    \mathcal{L}^{init}_{h} &= \lVert s(\vec{q}; \bm{\Psi}_h) - SDF(\vec{q}) \rVert^2_2 \\
    \mathcal{L}^{init}_{o} &= \lVert s(\vec{q}; \bm{\Psi}_o) - SDF(\vec{q}) \rVert^2_2.
\end{align}

\subsection{Geometry Decomposition and Canonicalization}
\label{subsection:reconstructon}
Given an input scan, we decompose and canonicalize the scan into two separate geometries of human and object, $\mesh{M}^{c*}_{h}, \mesh{M}^{c*}_{o}$, which minimizes the following total loss:
\begin{align}
\label{loss_geo}
    \mathcal{L}_{geo} &= \lambda^{rec}_{h_{geo}}\mathcal{L}^{rec}_{h_{geo}} + \lambda^{rec}_{o_{geo}}\mathcal{L}^{rec}_{o_{geo}} \\ \nonumber
    &+ \lambda^{SDS}_{h_{geo}}\mathcal{L}^{SDS}_{h_{geo}} + \lambda^{SDS}_{o_{geo}}\mathcal{L}^{SDS}_{o_{geo}} + \lambda^{seg}_{comp}\mathcal{L}^{seg}_{comp},
\end{align}
\begin{equation}
    \mesh{M}^{c*}_{h}, \mesh{M}^{c*}_{o} = \argmin_{\mesh{M}^{c}_{h}, \mesh{M}^{c}_{o}}\mathcal{L}_{geo}.
\end{equation}
We describe each loss in the following.

\begin{figure}[t]
\centering
\includegraphics[width=1.0\columnwidth, trim={0cm 0cm 0.5cm 0.4cm}]{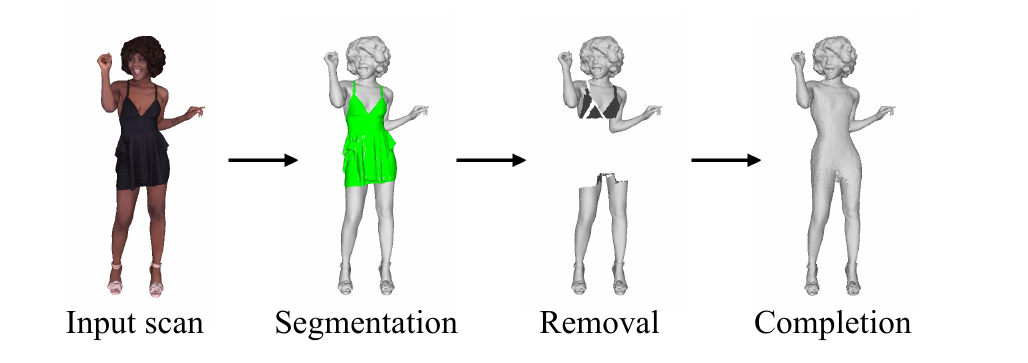}
\caption{\textbf{Decomposition and Synthesis.} We decompose humans and objects using 3D segmentation lifted from 2D and synthesize plausible geometry of the missing regions using pose-guided SDS.}
\label{fig:sds_geo}
\end{figure}

\paragraph{Reconstruction Loss.}
To decouple the geometry of the human and object, we employ the 3D surface segmentation of the target object. Specifically, we rasterize the scan from multiple viewpoints and perform binary segmentation in 2D, distinguishing the target object from other parts using an off-the-shelf open-vocabulary segmentation tool~\cite{kirillov2023sam}. Utilizing the aggregated pixel-to-face correspondence established during the rasterization process, we cast votes for each face of the mesh to determine whether it belongs to the specified object or not. Consequently, the given input scan in posed space, denoted as $\mesh{M}^{scan}$, is partitioned into two incomplete surface meshes: the object mesh, $\mesh{M}^{scan}_o$, and the remaining human figure mesh, $\mesh{M}^{scan}_h$, as shown in ~\cref{fig:sds_geo}.

To preserve the identity of visible regions of the input scan, we employ rendering-based reconstruction losses in the posed space. Using a differentiable rasterizer $\mathcal{R}$ and a sampled camera $\vec{k}$, we render masks $\mat{A}\in\{0, 1\}^{H \times W}$ and normal maps $\mat{N}\in\mathbb{R}^{H \times W}$ of the generated posed meshes $\mesh{M}^p_h$ and $\mesh{M}^p_o$, where $H, W$ are the height and width of the rendered masks and normal maps.
\begin{align}
\label{eq:render_recon}
\mat{A}^p_h, \mat{N}^p_h &= \mathcal{R}(\mesh{M}^p_h, \vec{k})
                          = \mathcal{R}(LBS(\mesh{M}^c_h), \vec{k}) \\
\mat{A}^p_o, \mat{N}^p_o &= \mathcal{R}(\mesh{M}^p_o, \vec{k})
                          = \mathcal{R}(LBS(\mesh{M}^c_o), \vec{k})
\end{align}
Together with the mask and normal map of the input mesh, we additionally render segmentation masks $\mat{S}^{scan}_h, \mat{S}^{scan}_o\in\{0, 1\}^{H \times W}$ for the human and the object using the 3D surface segmentation:
\begin{align}
\label{eq:render_recon}
\mat{A}^{scan}, \mat{N}^{scan}, \mat{S}^{scan}_h, \mat{S}^{scan}_o = \mathcal{R}(\mesh{M}^{scan}, \vec{k}).
\end{align}
Finally, the losses for reconstruction are defined as follows:
\begin{align}
\label{eq:recon}
\mathcal{L}^{rec}_{h_{geo}} &= \lVert \mat{N}^p_h \odot \mat{S}^{scan}_h - \mat{N}^{scan} \odot \mat{S}^{scan}_h\rVert^2_2, \\
\mathcal{L}^{rec}_{o_{geo}} &= \lVert \mat{N}^p_o \odot \mat{S}^{scan}_o - \mat{N}^{scan} \odot \mat{S}^{scan}_o\rVert^2_2 \\ 
                        &+ \lVert \mat{A}^p_o - \mat{A}^{scan} \odot \mat{S}^{scan}_o \rVert^2_2, \nonumber
\end{align}
where $\odot$ is the Hadamard product.
We employ extra mask loss to regularize the shape of the object in posed space, assuming that the object is layered on top of the human.
Furthermore, to capture the intricate details of human faces and hands, we render close-up views of these regions by zooming in on the corresponding joints of the posed SMPL-X mesh and apply the same reconstruction losses.

\paragraph{Pose-guided SDS Loss.}
Our goal is to obtain complete 3D assets in a neutral pose from a \emph{single} posed scan, which can then be animated into arbitrary poses without undesirable artifacts. 
The core challenges lie in the difficulty of (1) completing the occluded regions of both assets and (2) modeling canonical shape of each asset from a single scan.
To overcome both challenges,
we propose a pose-guided SDS loss that leverages the prior of the pretrained diffusion model equipped with ControlNet~\cite{zhang2023controlnet} conditioned with OpenPose poses~\cite{cao2019openpose}.
The gradient of our pose-guided SDS loss is defined as:
\begin{dmath}
\label{eq:pose_sds}
    \nabla_{\bm{\Psi}} \mathcal{L}^{SDS}_{pose}(\vec{z}_t(\mat{X}), \vec{y}, \vec{p}, \bm{\phi}) \\
    =\mathop{\mathbb{E}} [\omega(t)(\hat{\epsilon_{\bm{\phi}}}(\vec{z}_t(\mat{X}); \vec{y}, \vec{p}, t) - \epsilon) \frac{\partial \mat{X}}{\partial \bm{\Psi}} \frac{\partial \vec{z}_t(\mat{X})}{\partial \mat{X}}],
\end{dmath}
where $\mat{X}$ is the rendered normal or texture of the mesh $\mesh{M}$, $\vec{z}_t(\mat{X})$ is the latent embedding with noise from the forward process. $\vec{y}$ represents the positive and negative text prompts where positive prompts describe the underlying human and negative prompts describe the target object to remove. $\vec{p}$ is the pose condition for ControlNet~\cite{zhang2023controlnet} converted by the mapping from SMPL-X joints to OpenPose joints.

However, when the pose-guided SDS loss and reconstruction loss are applied in the posed space through the forward transformation of \cref{eq:warp}, the output canonical shape suffers from undesired artifacts due to many-to-one mapping from the canonical space to the posed space (see \cref{fig:ablation_qual} (b)). While previous approaches~\cite{chen2021snarf, Weng_2022_CVPR} address this ambiguity by jointly learning from multiple scans or images with various poses, we observe that these approaches perform poorly when given only a single scan.

To enable plausible canonicalization from a single scan, 
we apply our pose-guided SDS loss (\cref{eq:pose_sds}) in the canonical space. The gradients are derived as follows:
\begin{align}
    \nabla_{\bm{\Psi}_h}\mathcal{L}^{SDS}_{h_{geo}}&=\nabla_{\bm{\Psi}_h}\mathcal{L}^{SDS}_{pose}(\vec{z}_t(\tilde{\mat{N}}^{c}_{h}), \vec{y}_{h}, \vec{p}^{c}, \bm{\phi}), \label{eq:sds_geo_hum} \\
    \nabla_{\bm{\Psi}_o}\mathcal{L}^{SDS}_{o_{geo}}&=\nabla_{\bm{\Psi}_o}\mathcal{L}^{SDS}_{pose}(\vec{z}_t(\tilde{\mat{N}}^{c}_{comp}), \vec{y}_{comp}, \vec{p}^{c}, \bm{\phi}), \label{eq:sds_geo_obj}
\end{align}
where $\mathcal{L}^{SDS}_{h_{geo}}, \mathcal{L}^{SDS}_{o_{geo}}$ are the loss for the human and object space, respectively. $\tilde{\mat{N}}^{c}_{h}$, $\tilde{\mat{N}}^{c}_{comp}$ are the rendered normal map concatenated with the mask of the human mesh and the composite mesh in the canonical space, and $\vec{z}_t(\tilde{\mat{N}}^{c}_{h})$, $\vec{z}_t(\tilde{\mat{N}}^{c}_{comp})$ are the downsampled version of them with noise produced by the forward diffusion process as in Fantasia3D~\cite{chen2023fantasia3d}. $\vec{y}_{h}$, $\vec{y}_{comp}$ are the text prompts for the human and object space, and $\vec{p}^{c}$ is the neutral pose condition.
Remarkably, our pose-guided SDS loss in the canonical space along with the reconstruction loss in the posed space, effectively inpaints the occluded regions and eliminates the artifacts caused by the many-to-one mapping between the canonical space and posed space. To further remove the artifacts tightly attached to the human torso and assure the quality of decomposition in the input pose, we additionally apply our pose-guided SDS loss with a set of pre-defined poses including the input pose. 

For the object space, we apply our pose-guided SDS loss to the canonical composite mesh $\mesh{M}_{comp}^c$ (\cref{eq:sds_geo_obj}) with the gradient of the human mesh detached. Since the OpenPose ControlNet~\cite{zhang2023controlnet} is trained to generate pose-consistent human images, we obtain better guidance for the object space through pose-guided SDS loss with the rendering of the composite mesh than the object mesh. Please refer to the supplementary material for details.

\paragraph{Segmentation Loss}
The aforementioned reconstruction loss constrains each layer in isolation. However,
we observe that this alone is not sufficient to prevent penetration of the layer beneath when incomplete regions are synthesized via pose-guided SDS loss. Thus, we additionally incorporate a segmentation loss to further regularize the geometry after composition. Specifically, we assign one-hot encoded vector attributes $[1, 0]$ and $[0, 1]$, respectively to every face of $\mesh{M}^p_h$ and $\mesh{M}^p_o$, and rasterize both meshes together to get the segmentation masks for the human and the object, $\mat{S}^p_{h}$ and $\mat{S}^p_{o}$. We minimize the difference between $\mat{S}^p_{h}$ and $\mat{S}^p_{o}$, and the rendered segmentation masks of the input scan, $\mat{S}^{scan}_h$ and $\mat{S}^{scan}_o$, with the following loss:
\begin{align}
\label{eq:segmentation}
\mathcal{L}^{seg}_{comp} &= \lVert \mat{S}^p_{h}  - \mat{S}^{scan}_h \rVert^2_2
                          + \lVert \mat{S}^p_{o}  - \mat{S}^{scan}_o \rVert^2_2.
\end{align}

\begin{figure}[t]
\includegraphics[width=1.0\columnwidth, trim={0.6cm 0cm 0.2cm 0cm}]{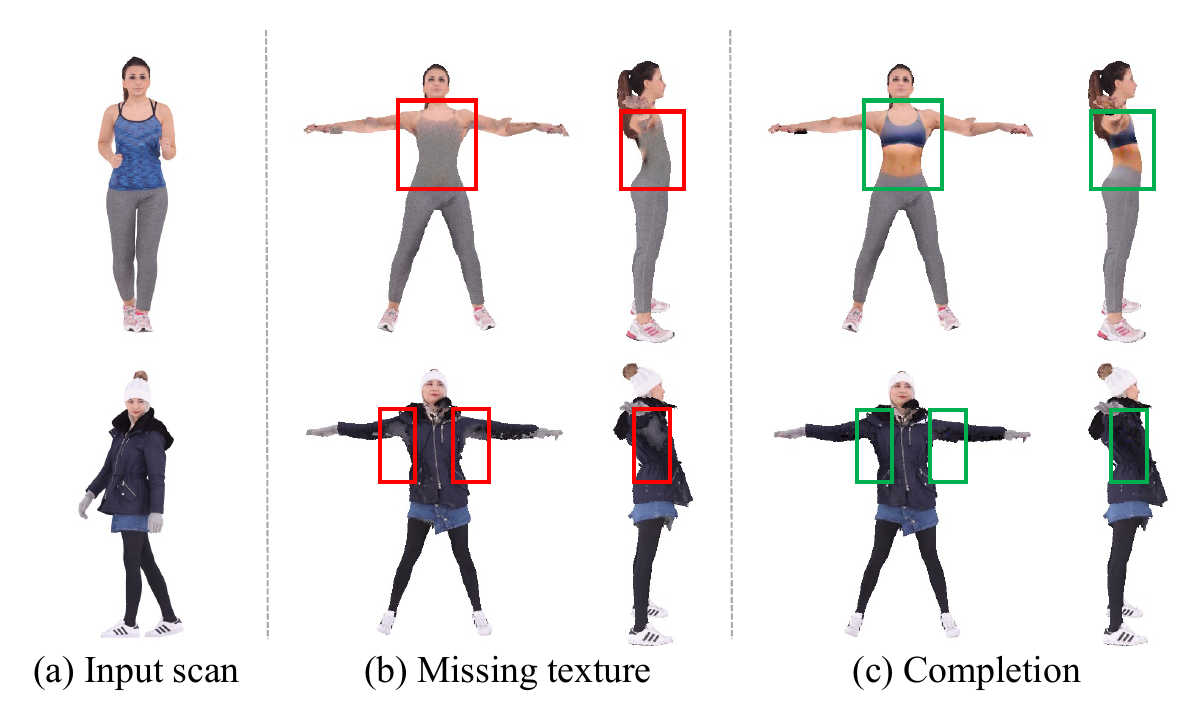}
\caption{\textbf{Texture Generation.} Applying SDS loss in canonical space generates texture for regions occluded by objects along with self-occluded regions.}
\label{fig:sds_tex}
\end{figure}

\subsection{Appearance Completion}
\label{subsection:appearance}
Given the inpainted canonical human mesh $\mesh{M}^c_h$, and object mesh $\mesh{M}^c_o$, we model the appearance of each mesh represented as vertex colors. We employ MLP networks $\bm{\Gamma}_h$ and $\bm{\Gamma}_o$ to predict the albedo of every vertex. 

The total loss for optimizing the texture is defined as,
\begin{align}
\label{loss_tex}
    \mathcal{L}_{tex} &= \lambda^{rec}_{h_{tex}}\mathcal{L}^{rec}_{h_{tex}} + \lambda^{rec}_{o_{tex}}\mathcal{L}^{rec}_{o_{tex}}\nonumber \\ &+ \lambda^{SDS}_{h_{tex}}\mathcal{L}^{SDS}_{h_{tex}} + \lambda^{SDS}_{o_{tex}}\mathcal{L}^{SDS}_{o_{tex}}.
\end{align}
Similar to \cref{subsection:reconstructon}, we utilize the 3D surface segmentation to initialize the color of the visible regions in the input mesh. Specifically, we differentiably render RGB images, $\mat{I}_h^p$, $\mat{I}_o^p$, and $\mat{I}^{scan}$ of the posed meshes, $\mesh{M}^p_h$ and $\mesh{M}^p_o$, and the input scan, $\mesh{M}^{scan}$, and optimize the following losses:
\begin{align}
\label{eq:recon_tex}
\mathcal{L}^{rec}_{h_{tex}} &= \lVert \mat{I}^p_h \odot \mat{S}^{scan}_h - \mat{I}^{scan} \odot \mat{S}^{scan}_h\rVert^2_2, \\
\mathcal{L}^{rec}_{o_{tex}} &= \lVert \mat{I}^p_o \odot \mat{S}^{scan}_o - \mat{I}^{scan} \odot \mat{S}^{scan}_o\rVert^2_2.
\end{align}

To generate textures for the fully occluded regions, we utilize the pose-guided SDS loss as shown in ~\cref{fig:sds_tex}. 
We use the vertex colors of $\mesh{M}^c_h$ and $\mesh{M}^c_{comp}$ to render the RGB images, $\mat{I}^c_h$ and $\mat{I}^c_{comp}$, and optimize our texture MLPs, $\bm{\Gamma}_h$ and $\bm{\Gamma}_o$, by computing the gradients of following pose-guided SDS losses:
\begin{align}
    \nabla_{\bm{\Gamma}_h}\mathcal{L}^{SDS}_{h_{tex}}&=\nabla_{\bm{\Gamma}_h}\mathcal{L}^{SDS}_{pose}(\vec{z}_t(\mat{I}^{c}_{h}), \vec{y}_{h}, \vec{p}^{c}, \bm{\phi}), \label{eq:sds_tex_hum} \\
    \nabla_{\bm{\Gamma}_o}\mathcal{L}^{SDS}_{o_{tex}}&=\nabla_{\bm{\Gamma}_o}\mathcal{L}^{SDS}_{pose}(\vec{z}_t(\mat{I}^{c}_{comp}), \vec{y}_{comp}, \vec{p}^{c}, \bm{\phi}), \label{eq:sds_tex_obj}
\end{align}
where $\vec{z}_t(\mat{I}^{c}_{h})$ and $\vec{z}_t(\mat{I}^{c}_{comp})$ represent the latent embeddings of $\mat{I}^c_h$ and $\mat{I}^c_{comp}$, achieved using the pretrained image encoder of the diffusion model~\cite{rombach2022latent}. All other notations remain consistent with those used in \cref{eq:sds_geo_hum} and \cref{eq:sds_geo_obj}.

\subsection{Composition}
\label{subsection:composition}

When composing the generated assets to novel identities, penetration of the human layer beneath could happen. To resolve this, we also introduce a refinement step. Given a canonical human mesh $\mesh{M}^c_h$, and a canonical object mesh $\mesh{M}^c_o$, we optimize the vertex positions of $\mesh{M}^c_h$ along their normal directions, $\vec{n}_h$. For each vertex $\vec{v}_h \in \mathcal{V}^c_h$ of $\mesh{M}^c_h$, we find its nearest neighbor vertex $\vec{v}^{nn}_h \in \mathcal{V}^{c'}_o \in \mathcal{V}^c_o$ of $\mesh{M}^c_o$, where $\mathcal{V}^{c'}_o$ denotes the visible vertices among $\mathcal{V}^c_o$. We introduce a penalty when $\overrightarrow{\vec{v}_h\vec{v}^{nn}_h}$ and $\vec{n}_h$ are oriented in opposite directions. Similarly, for each vertex $\vec{v}_o \in \mathcal{V}^{c'}_o$, we find its nearest neighbor vertex $\vec{v}^{nn}_o \in \mathcal{V}^c_h$, and penalize when $\overrightarrow{\vec{v}_o\vec{v}^{nn}_o}$ and $\vec{n}_o$ have the same direction, where $\vec{n}_o$ denotes the normals of $\vec{v}_o$. Formally, we minimize the following loss,
\begin{align}
\label{loss_refine}
    \mathcal{L}_{ref} &= - \frac{\overrightarrow{\vec{v}_h\vec{v}^{nn}_h}}{\lVert \overrightarrow{\vec{v}_h\vec{v}^{nn}_h} \rVert} \cdot \vec{n}_h + \frac{\overrightarrow{\vec{v}_o\vec{v}^{nn}_o}}{\lVert \overrightarrow{\vec{v}_o\vec{v}^{nn}_o} \rVert} \cdot \vec{n}_o + \lambda_{dis} \lVert \Delta \vec{v}_h \rVert^2_2,
\end{align}
where the last term regularizes the displacements of $\vec{v}_h$.

\section{Experiments}
\label{sec:experiments}

\subsection{Datasets and Metrics}

\noindent \textbf{RenderPeople~\cite{renderpeople}:} RenderPeople provides high-quality sigle-layer 3D human scans, and we select 30 scans to cover diverse categories of target objects to decompose. We evaluate the quality of the decomposition against state-of-the-art (SOTA) methods~\cite{brooks2022instructpix2pix, sella2023voxe}. Following the evaluation protocol of previous editing work~\cite{gal2022stylegannada, brooks2022instructpix2pix, hqaue2023in2n}, we utilize the CLIP text-image direction similarity which measures the alignment of the performed edit with the text instruction. We also present a novel metric named, pixel-wise object removal score (POR Score), which measures the ratio of the number of pixels of the target object, before and after the edit. During evaluation, we render both the input and the edited output from 30 evenly distributed viewpoints and measure each metric.

\noindent \textbf{CAPE Dataset}~\cite{ma2020cape} CAPE dataset contains the 3D sequences of clothed humans along with the corresponding SMPL parameters. We utilize CAPE dataset to evaluate the quality of canonicalization in comparison to existing methods and conduct ablation studies. For evaluation, we use 18 subjects, each wearing diverse clothing types that include both long and short upper and lower garments. For each subject, we select 100 scans with equal intervals in the sequence, and perform canonicalization using the last scan. We then pose the modeled canonical shape into poses of the preceding 99 scans, and calculate Intersection over Union (IoU) and Chamfer distance (Chamf) to measure the alignment. Since the dataset provides parameters of SMPL, we adapt our pipeline to use SMPL instead of SMPL-X.\footnote{All datasets used in this research were exclusively downloaded, accessed, and utilized at SNU.}

\subsection{Qualitative Evaluation}
\label{sec:qualitative}

\begin{figure*}[ht]
\centering
\includegraphics[width=1.0\linewidth, trim={0cm 0 0cm 0.5cm}]{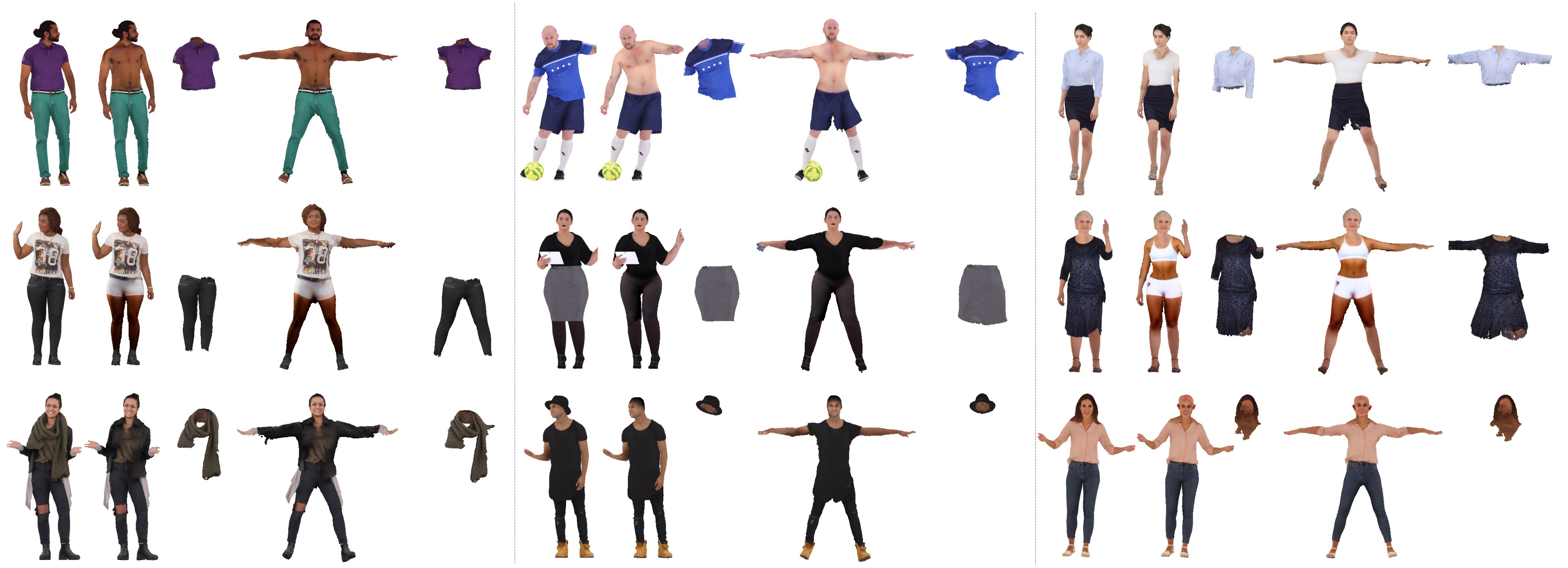}
\vspace{-15px}
\caption{\textbf{Decomposition and Canonicalization.} In each set, we show the decomposition and canonicalization results of the leftmost sample.}
\label{fig:decomposition}
\vspace{-10px}
\end{figure*}

\noindent \textbf{Decomposition and Canonicalization.}
\cref{fig:decomposition} shows that our method synthesizes realistic geometry and texture for the occluded regions, and enables robust canonicalization of both humans and objects, even in challenging poses.

\noindent \textbf{Layered Decomposition.}
In \cref{fig:teaser}, we highlight the key advantage of our method by applying a series of decompositions to the input scan. 
By recomposing the decomposed assets, our method enables the decomposition of specific layers of clothing which was previously not feasible.

\begin{figure}[t]
\includegraphics[width=1.0\columnwidth, trim={0.4cm 0 0.6cm 0.5cm}]{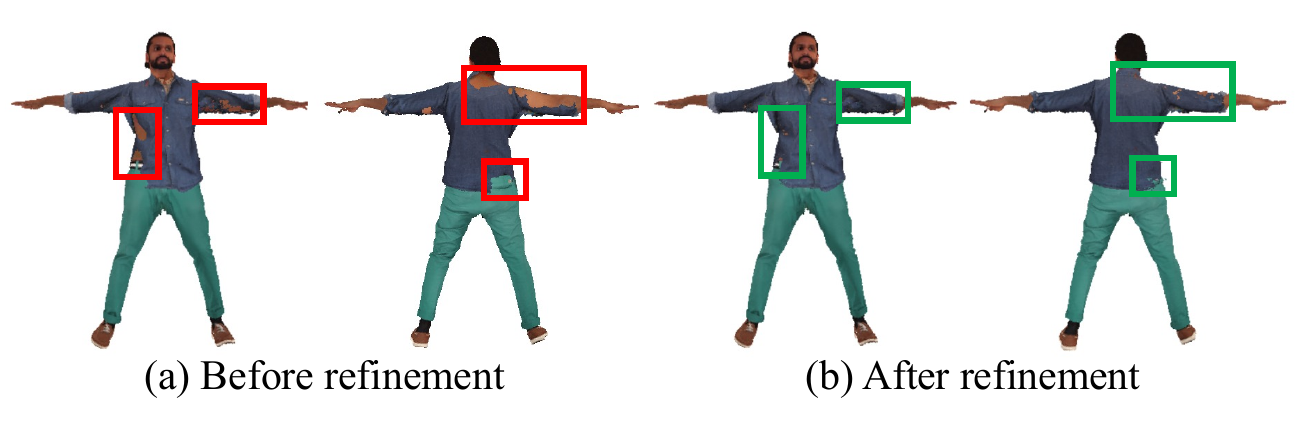}
\vspace{-17px}
\caption{\textbf{Refinement.} Our refinement stage successfully reduces the misalignment between humans and objects.}
\label{fig:refinement}
\vspace{-7px}
\end{figure}

\noindent \textbf{Composition and Refinement.}
\cref{fig:teaser} shows that our method enables avatar customization with various combinations of the decomposed assets. The composition outputs can be further refined as shown in Fig.~\ref{fig:refinement}.

\begin{figure}[t]
\includegraphics[width=1.0\columnwidth, trim={2cm 0 0.2cm 0.8cm}]{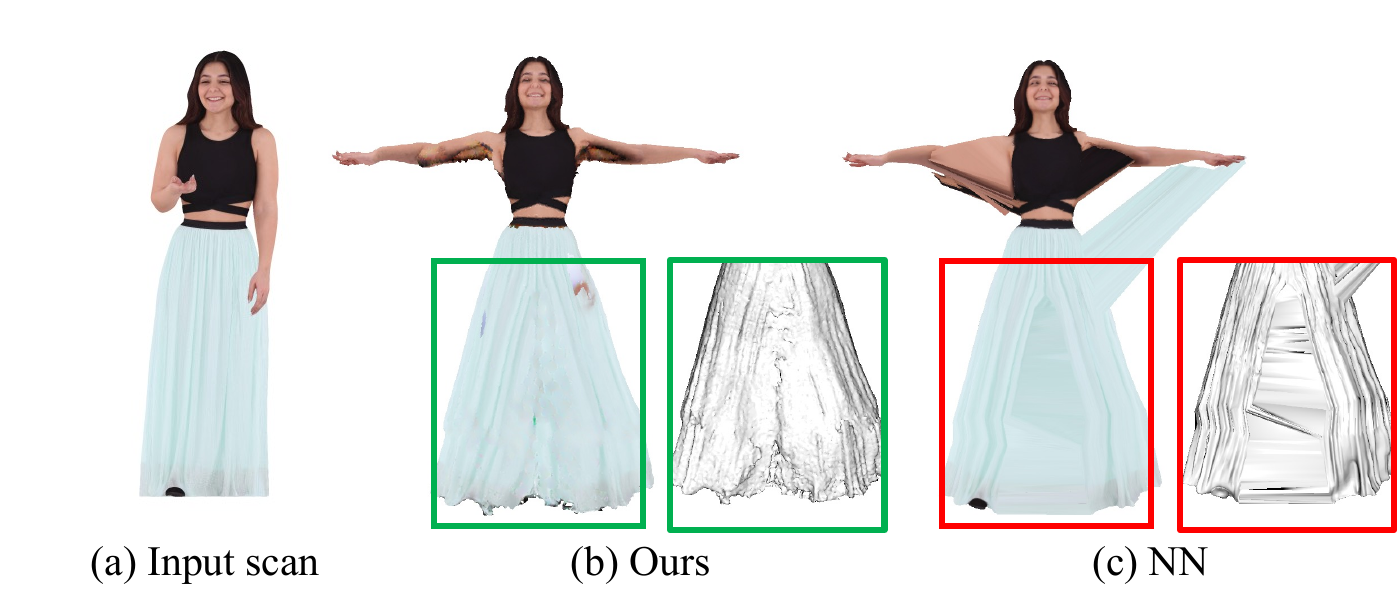}
\vspace{-17px}
\caption{\textbf{Loose Clothing.} Our method excels in modeling the canonical geometry of loose clothing such as dresses or skirts compared to existing canonicalization methods.}
\label{fig:loose_clothing}
\vspace{-14px}
\end{figure}

\noindent \textbf{Loose Clothing.}
As shown in \cref{fig:loose_clothing}, our approach enables the successful canonicalization and modeling of loose clothing, where a simple canonicalization method based on nearest neighbor~\cite{huang2020arch,he2021arch++} struggles.

\begin{table}[t]

\centering
\small{

\begin{tabular}{lccc}
\toprule
       & CLIP TI Direction Similarity $\uparrow$ & POR Score $\downarrow$ \\
\midrule
Ours & \textbf{0.1117} & \textbf{0.1144} \\
I-N2N~\cite{hqaue2023in2n} & 0.0621 & 0.4871 \\
Vox-E~\cite{sella2023voxe} & 0.0374 & 0.5583 \\
\bottomrule
\end{tabular}
}
\caption{\textbf{Quantitative comparison on decomposition.} We report CLIP similarity and pixel-wise object removal score to provide quantitative metrics for the subjective editing task.}

\label{tab:decomp_clip_quant}
\end{table}

\begin{figure}
    \centering
    \rotatebox{90}{\quad \qquad (-) hair \quad \qquad (-) hat \qquad \quad (-) pants}
    \begin{subfigure}[b]{0.1\textwidth}
        \centering
        \includegraphics[trim=0 0 0 500, clip, width=\textwidth]{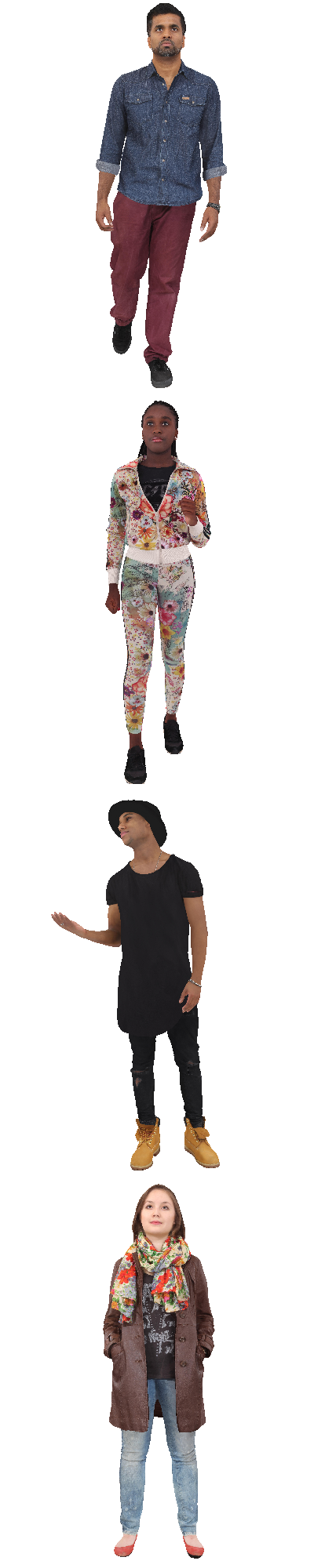}
        \caption{Input}
        \label{fig:qual_gt}
    \end{subfigure}
    \begin{subfigure}[b]{0.1\textwidth}
        \centering
        \includegraphics[trim=0 0 0 500, clip, width=\textwidth]{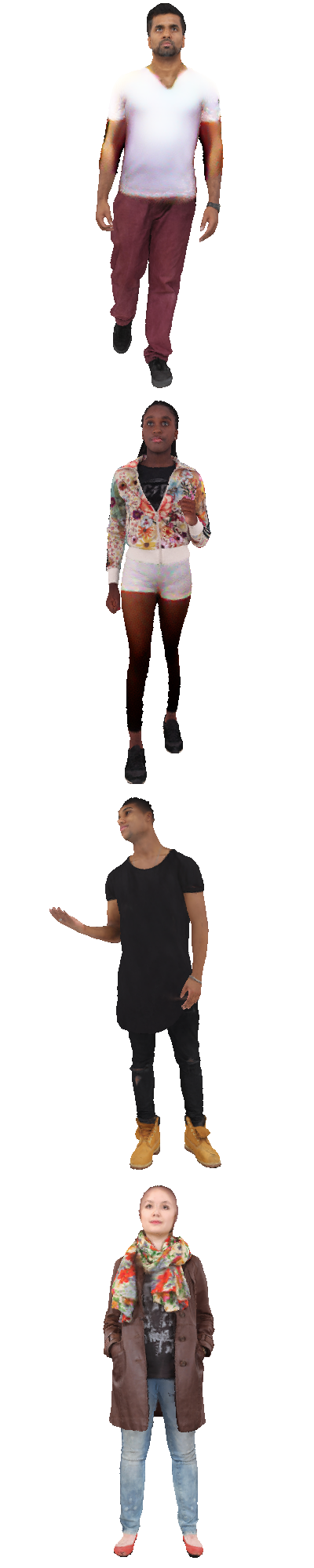}
        \caption{Ours}
        \label{fig:qual_ours}
    \end{subfigure}
    \begin{subfigure}[b]{0.1\textwidth}
        \centering
        \includegraphics[trim=0 0 0 500, clip, width=\textwidth]{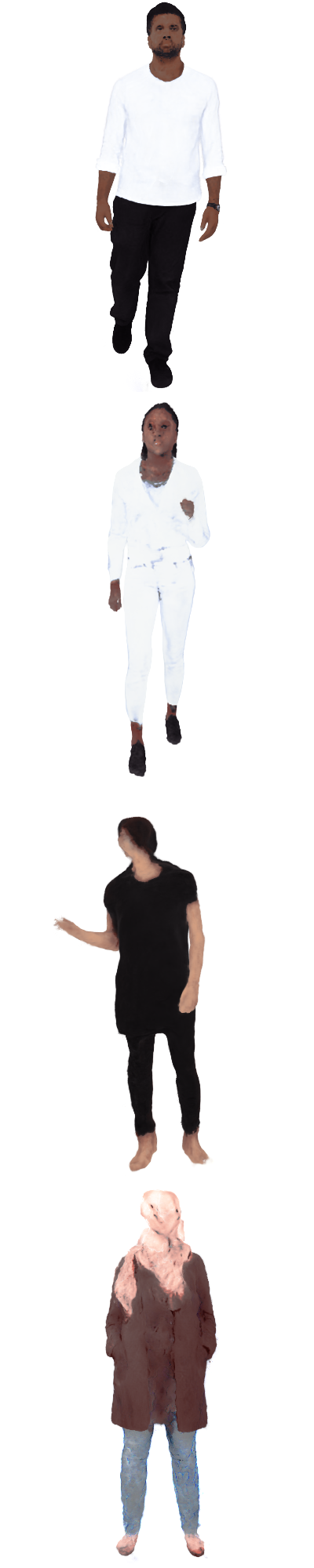}
        \caption{I-N2N~\cite{hqaue2023in2n}}
        \label{fig:qual_in2n}
    \end{subfigure}
    \begin{subfigure}[b]{0.1\textwidth}
        \centering
        \includegraphics[trim=0 0 0 500, clip, width=\textwidth]{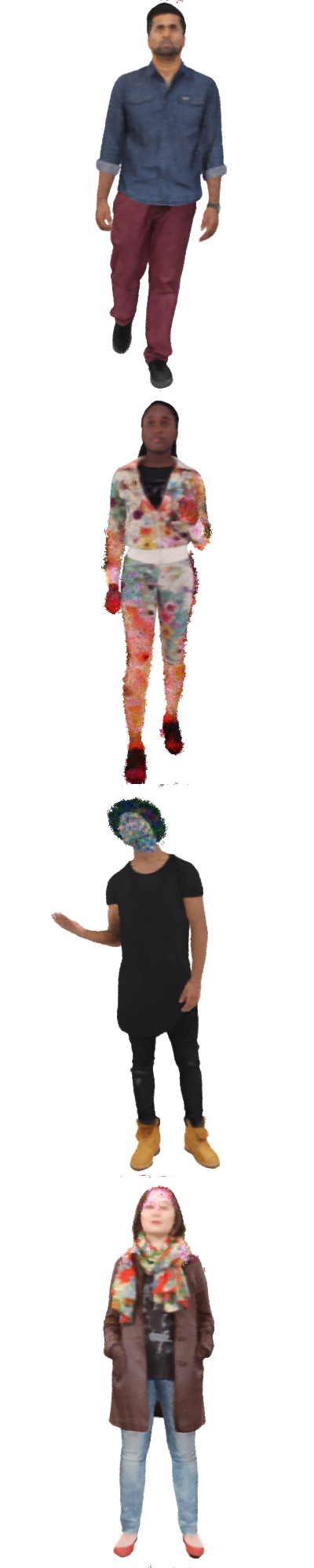}
        \caption{Vox-E~\cite{sella2023voxe}}
        \label{fig:qual_vox-e}
    \end{subfigure}     
    \caption{\textbf{Qualitative Comparison.} In contrast to our approach, other methods often face challenges in effectively removing the intended object or resulting in deterioration in unrelated areas.}
    \label{fig:decomp_clip_qual}
    \vspace{-10px}
\end{figure}

\subsection{Quantitative Evaluation}
\label{sec:quantitative}
\noindent \textbf{Decomposition.} We evaluate the quality of decomposition against the SOTA text-guided 3D editing methods~\cite{hqaue2023in2n, sella2023voxe}, which we believe is the closest to our task. Instruct-NeRF2NeRF~\cite{hqaue2023in2n} is a text-guided NeRF~\cite{mildenhall2020nerf} editing method based on the Instruct-Pix2Pix~\cite{brooks2022instructpix2pix}. Vox-E~\cite{sella2023voxe} edits a 3D scene by first fitting a ReLU field~\cite{karnewar2022relu} with multi-view images and then editing the learned ReLU field using SDS loss. We provide prompts for each method to remove the target object and compare the decomposition results. \cref{tab:decomp_clip_quant} shows that our method outperforms SOTA baselines, achieving the highest CLIP text-to-image similarity and the lowest POR Score. We also provide qualitative comparison in \cref{fig:decomp_clip_qual}.

\begin{figure}[t]
\includegraphics[width=1.0\columnwidth, trim={0.5cm 0cm 0.5cm 0.0cm}]{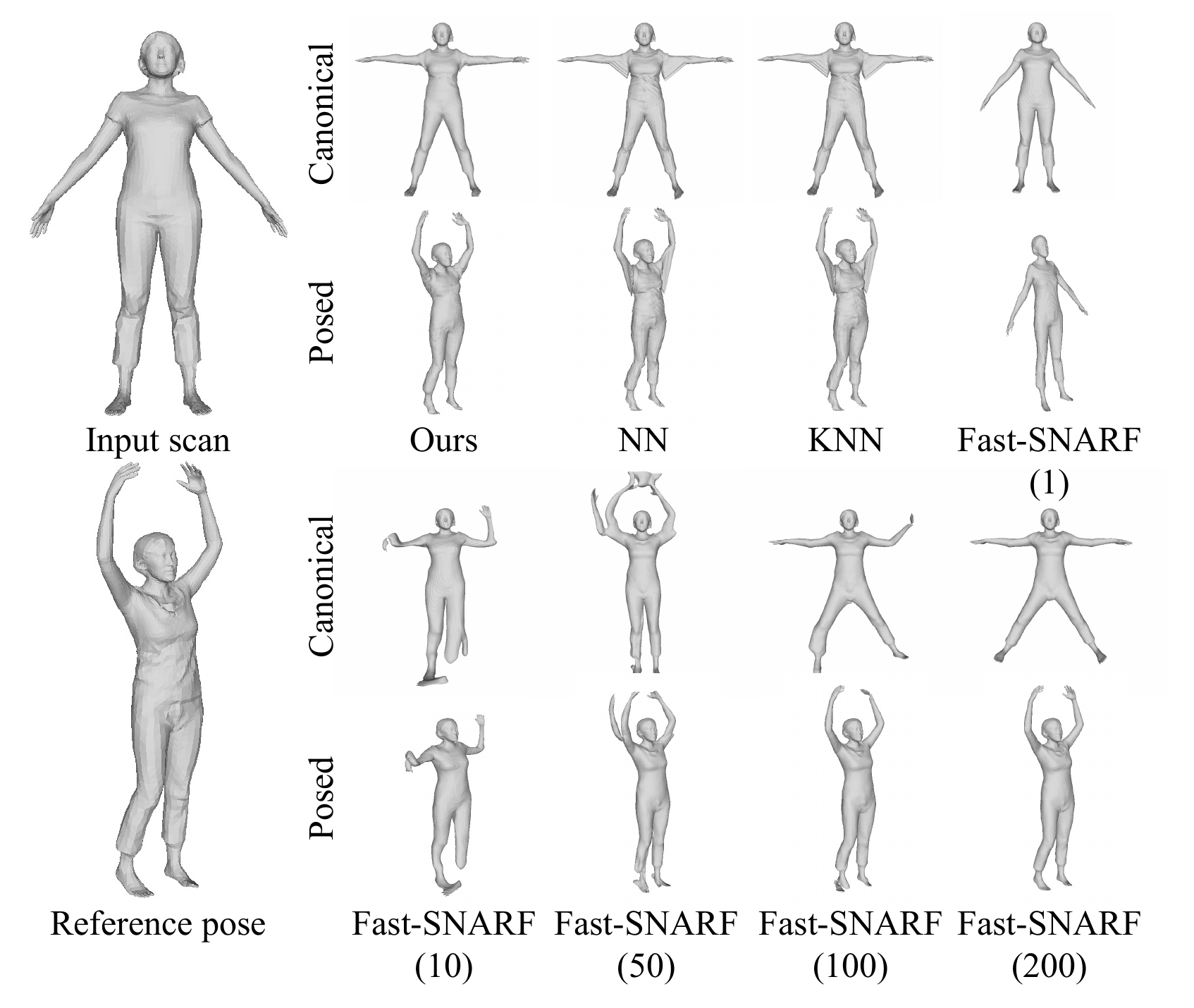}
\vspace{-10px}
\caption{\textbf{Qualitative comparison on canonicalization.} 
We present the results of single-scan canonicalization in the top two rows. The bottom two rows depict the results of Fast-SNARF~\cite{chen2023fast}, with varying numbers of training scans denoted in the parenthesis.}
\label{fig:cano_qual}
\end{figure}

\begin{table}[t]

\centering
\small{
\begin{tabular}{lcc}
\toprule
Method & IoU$\uparrow$ & Chamf$\downarrow$ \\
\midrule
Ours & 84.70\% & 0.821 \\
NN & 83.93\% & 0.845 \\
KNN & 83.93\% & 0.846 \\
Fast-SNARF (w/ 1 scan)~\cite{chen2023fast} & 38.97\% & 6.778 \\
Fast-SNARF (w/ 10 scans) & 67.45\% & 3.029 \\
Fast-SNARF (w/ 50 scans) & 81.11\% &  1.430 \\
Fast-SNARF (w/ 100 scans) & 94.01\% &  0.435 \\
Fast-SNARF (w/ 200 scans) & 96.55\% &  0.315 \\
\bottomrule
\end{tabular}
}
\caption{\textbf{Quantitative comparison of canonicalization.} Chamfer distances are in centimeters. We use $K=6$ for KNN.}
\vspace{-10px}
\label{tab:cano_quant}
\end{table}

\noindent \textbf{Canonicalization.}
We compare our canonicalization results with baseline methods. 
To solely assess the quality of canonicalization, we exclude the decomposition process by modeling the whole scan in a single space.
We employ three baselines for comparison. Nearest Neighbor (NN), transforms each vertex to its canonical position based on the skinning weights of the nearest neighbor SMPL vertex~\cite{huang2020arch}. K-Nearest Neighbor (KNN) uses the weighted average of skinning weights of k-nearest neighbor SMPL vertices~\cite{yang2018analyzing}. \cref{tab:cano_quant} demonstrates that our method outperforms the baselines, reporting the highest IoU and the lowest Chamfer distance when transformed into various poses.
We also compare our results with Fast-SNARF~\cite{chen2023fast}, the current SOTA for canonicalization from multiple scans. However, we observed severe instability in the learning of MLP-based skinning fields with a small number of scans. Thus, we discard the skinning field in Fast-SNARF, and use the nearest neighbor skinning weights instead for comparison. \cref{tab:cano_quant} shows that our method outperforms Fast-SNARF trained with up to 50 scans. Note that the original Fast-SNARF is trained with a significantly larger dataset of around 3000 scans. The qualitative comparison is presented in \cref{fig:cano_qual}.

\begin{table}[t]
\centering

\small{
\begin{tabular}{cccc}
\toprule
Cano. SDS Loss & Pose-Guided SDS  & IoU$\uparrow$ & Chamf$\downarrow$ \\
\midrule
\xmark & \xmark & 79.97\% & 1.384 \\
\cmark & \xmark & 82.89\% & 1.227 \\
\cmark & \cmark & \textbf{83.59\%} & \textbf{1.184} \\
\bottomrule
\end{tabular}
}
\caption{
\textbf{Ablation study.} We ablate the SDS loss in the canonical space and the pose-guided SDS loss.
}
\vspace{-5px}
\label{tab:ablation_quant}
\end{table}

\begin{figure}[t]
\includegraphics[width=1.0\columnwidth, trim={1.2cm 0cm 0cm 0.0cm}]{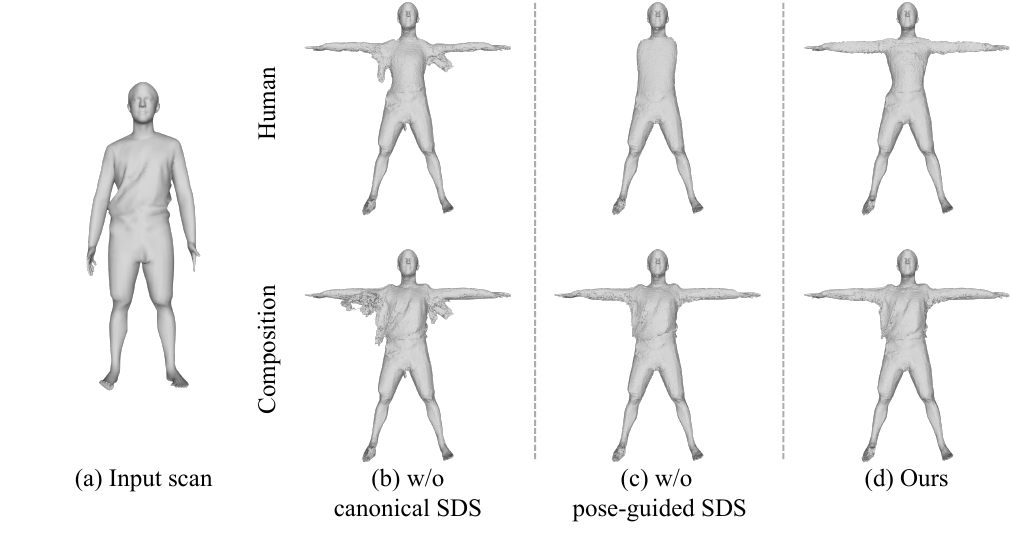}
\caption{\textbf{Ablation study.} We show the effect of applying SDS loss in canonical space and the importance of the pose-guided SDS loss for robust canonicalization.}
\label{fig:ablation_qual}
\end{figure}

\noindent \textbf{Ablation Study.}
\cref{tab:ablation_quant} and \cref{fig:ablation_qual} summarize an ablation study to evaluate our design choices. 
First, we validate the importance of the SDS loss in the canonical space. Without the SDS loss in the canonical space, we observe artifacts in the canonical shape as shown in \cref{fig:ablation_qual} (b), leading to implausible reposing results. We further validate our pose-guided SDS by using the vanilla SDS loss without a pose condition. As illustrated in \cref{fig:ablation_qual} (c), the use of the vanilla SDS loss leads to noticeable artifacts near the armpits and often lack large body parts. In contrast, using the proposed pose-guided SDS loss achieves more plausible canonicalization without artifacts as shown in \cref{fig:ablation_qual} (d) and \cref{tab:ablation_quant}.

\section{Discussion and Future Work}
\label{sec:conclusion}

We presented GALA, a framework that turns a single static scan into reusable and animatable layered assets. Our experiments show that decomposing and inpainting separated layers in 3D is now possible with the help of a powerful 2D diffusion prior. The proposed pose-guided SDS loss allows us to jointly optimize each component in both posed and canonical space to produce clean textured 3D geometry.
The resulting layered assets can be composed with novel identities in a plausible manner and be further reposed to a target pose. 
We also demonstrate that our method outperforms existing editing methods both qualitatively and quantitatively.

\noindent\textbf{Limitation and Future Work.}
Our approach currently generates a static canonical shape for reposing. Modeling pose-dependent deformation of clothing from a single scan can be addressed in future work. 
The dependency on accurate 2D segmentation can be also problematic if the 2D segmentation module fails. Self-discovering each layer without requiring 2D segmentation is also an interesting future work.

\noindent\textbf{Acknowledgements:} The work of SNU members was supported by NRF grant funded by the Korean government (MSIT) (No. 2022R1A2C2092724), and IITP grant funded by the Korean government (MSIT) (No.2022-0-00156, No.2021-0-01343). H. Joo is the corresponding author.

{\small
\setlength{\bibsep}{0pt}
\bibliographystyle{abbrvnat}
\bibliography{shortstrings, 11_references}
}

\ifarxiv 
\clearpage
\appendix
\label{sec:appendix}

\section{Implementation Details}
\label{sec:implemention_details}

\subsection{Network Architectures}
\label{sec:network}
We implement our networks for predicting SDF and offsets, $\bm{\Psi}_h$ and $\bm{\Psi}_o$, as a 2-layer MLP network with 32 hidden units and ReLU activations except for the last layer. As inputs, each network takes the 3D Cartesian coordinates of the vertices, $X_T$, of the designated canonical DMTet grid, $(X_T, T)$. The coordinates are normalized between 0 to 1, and encoded using a hash positional encoding~ \cite{muller2022instant} with 16 resolution levels and a maximum resolution of 1024.
The networks for predicting vertex colors,  $\bm{\Gamma}_h$ and $\bm{\Gamma}_o$, are implemented using a 1-layer MLP network with 32 hidden units and ReLU activations except for the last layer that uses sigmoid activations. As inputs, each network takes the 3D Cartesian coordinates of the vertices of the canonical human mesh and object mesh, $\mesh{M}^c_h$ and $\mesh{M}^c_o$. The coordinates are similarly normalized between 0 to 1, and encoded using a hash positional encoding with 16 resolution levels and a maximum resolution of 2048.

\subsection{Optimization Details}
\label{sec:optimization}
The total loss, $\mathcal{L}_{geo}$, for geometry modeling is as follows:
\begin{align}
\label{loss_geo}
    \mathcal{L}_{geo} &= \lambda^{rec}_{h_{geo}}\mathcal{L}^{rec}_{h_{geo}} + \lambda^{rec}_{o_{geo}}\mathcal{L}^{rec}_{o_{geo}} + \lambda^{seg}_{comp}\mathcal{L}^{seg}_{comp}\\ \nonumber
    &+ \lambda^{SDS}_{h_{geo}}\mathcal{L}^{SDS}_{h_{geo}} + \lambda^{SDS}_{o_{geo}}\mathcal{L}^{SDS}_{o_{geo}},
\end{align}
where $\lambda^{rec}_{h_{geo}}=5\times10^3$, $\lambda^{rec}_{o_{geo}}=5\times10^3$, $\lambda^{seg}_{comp}=1\times10^5$, $\lambda^{SDS}_{h_{geo}}=1$, and $\lambda^{SDS}_{o_{geo}}=1$. We use AdamW optimizer with a learning rate of $0.001$ and optimize for 1600 steps, after 400 steps of the initialization process with $\mathcal{L}^{init}_{h}$ and $\mathcal{L}^{init}_{o}$.

The total loss, $\mathcal{L}_{tex}$, for appearance modeling is,
\begin{align}
\label{loss_tex}
    \mathcal{L}_{tex} &= \lambda^{rec}_{h_{tex}}\mathcal{L}^{rec}_{h_{tex}} + \lambda^{rec}_{o_{tex}}\mathcal{L}^{rec}_{o_{tex}}\nonumber \\ &+ \lambda^{SDS}_{h_{tex}}\mathcal{L}^{SDS}_{h_{tex}} + \lambda^{SDS}_{o_{tex}}\mathcal{L}^{SDS}_{o_{tex}},
\end{align}
where $\lambda^{rec}_{h_{tex}}=1\times10^8$ and $\lambda^{rec}_{o_{tex}}=1\times10^8$. $\lambda^{SDS}_{h_{tex}}=0$ and $\lambda^{SDS}_{o_{tex}}=0$ for the first 400 steps, and $\lambda^{SDS}_{h_{tex}}=1$ and $\lambda^{SDS}_{o_{tex}}=1$ otherwise. We use AdamW optimizer with a learning rate of $0.01$ and optimize for 2000 steps. Each stage takes about 20 minutes on a single NVIDIA RTX 3090.

\subsection{Additional Details}
\label{sec:additional}

\paragraph{Prompts for the SDS loss.} For $\vec{y}_{h}$ in $\nabla_{\bm{\Psi}_h}\mathcal{L}^{SDS}_{h_{geo}}$ and $\nabla_{\bm{\Gamma}_h}\mathcal{L}^{SDS}_{h_{tex}}$, we use ``A photo of a man/woman'' as the positive prompt and ``\emph{\{target object}\}'' as the negative prompt. Note that we use ``man'' or ``woman'' based on the gender provided by RenderPeople~ \cite{renderpeople} and CAPE Dataset~ \cite{ma2020cape}. For $\vec{y}_{comp}$ in $\nabla_{\bm{\Psi}_o}\mathcal{L}^{SDS}_{o_{geo}}$ and $\nabla_{\bm{\Gamma}_o}\mathcal{L}^{SDS}_{o_{tex}}$, we use ``A photo of a man/woman wearing \emph{\{target object}\}'' as the positive prompt and do not use any negative prompt. Following DreamFusion~ \cite{poole2023dreamfusion}, we incorporate view directions by concatenating ``front/side/back view'' to each prompt based on the viewing angle of the sampled camera.

\paragraph{Camera Sampling.} We set the camera center using spherical coordinate system, $(r, \theta, \phi)$, where $r$ denotes the radial distance from the origin, $\theta$ denotes the elevation, and $\phi$ denotes the azimuth angle. We set $r=3$, and sample cameras facing the origin from $\theta \in [-\frac{\pi}{18}, \frac{\pi}{9}]$, and $\phi \in [0, 2\pi]$. We also sample the field of view from $\mathcal{U}(\frac{\pi}{7}, \frac{\pi}{4})$. We additionally use zoomed-in views to capture fine details of human faces and hands and to effectively synthesize the missing regions where human and target object interact. To render zoomed-in images, we translate and scale the input mesh before the rendering process. For the zoomed-in views for faces and hands, we translate the input mesh using the corresponding joint information of the SMPL-X mesh such that each joint locates at the origin, and scale the input mesh by factor of 5 for rendering the face and 10 for rendering the hands. For the zoomed-in views for regions where human and target object interact, we utilize the bounding box information of the target object. Specifically, given the object bounding box $\vec{x}_{l}=(x_{min}, y_{min}, z_{min})$ to $\vec{x}_{r}=(x_{max}, y_{max}, z_{max})$, we first translate the input mesh by $t \sim \mathcal{U}(\frac{\vec{x}_{r}+3\vec{x}_{l}}{4}, \frac{3\vec{x}_{r}+\vec{x}_{l}}{4})$. We then scale the input mesh by the factor of $s \sim \mathcal{U}(\frac{1}{0.6 max(\vec{x}_{r} - \vec{x}_{l})}, \frac{1}{0.3max(\vec{x}_{r} - \vec{x}_{l})})$.

\section{Evaluation Details}
\label{sec:quant_details}

\subsection{Decomposition}
\paragraph{Baselines.} To the best of our knowledge, there is no existing work that tackles the decomposition of a 3D scan. Therefore, we use the recent text-based 3D editing methods as baseline: Instruct-NeRF2NeRF~ \cite{hqaue2023in2n} and Vox-E~ \cite{sella2023voxe}. For evaluation, we use the official implementation for both methods. We train nerfacto model~ \cite{mildenhall2020nerf} for Instruct-NeRF2NeRF and ReLU field~ \cite{karnewar2022relu} for Vox-E with each scan. Since Instruct-NeRF2NeRF is based on Instruct-Pix2Pix~ \cite{brooks2022instructpix2pix}, the prompt should be given in the form of ``instruction''; hence, the basic form of prompts we use for Instruct-NeRF2NeRF is ``Remove \emph{\{target object}\} from him/her'' or ``Change his/her \emph{\{target object}\} to a white t-shirt/shorts'' to avoid getting naked body for single-layered clothing. For Vox-E, the basic form of prompts we use is ``A photo of a man/woman without \emph{\{target object}\}''.

\paragraph{POR metric.} We propose a novel metric named pixel-wise object removal score (POR Score) for quantitatively evaluating the decomposition performance. Specifically, we render $30$ images per subject using the camera views with equally distributed yaw angles. Then, we run the off-the-shelf open-vocabulary image segmentation method, SAM~ \cite{kirillov2023sam}, to get the segmentation of the target object specified by the prompt. Ideally, if the target object is properly decomposed or removed, there should be no pixel classified as the target object for the images rendered after decomposition. Hence, we compute the ratio of the number of pixels classified as the target object in the images after editing and the images rendered from the input scan as follows:
\begin{equation}
    POR = \frac{1}{\lvert \mat{K} \rvert} \sum\limits_{\vec{k}\in\mat{K}}\frac{\sum\limits_{(i,j) \in \mat{M}_{\vec{k}}^{input}} \boldone(\mathrm{SAM}(\mat{I}_{\vec{k}}^{edit})_{ij}=1)}{\lvert \mat{M}_{\vec{k}}^{input} \rvert},
\end{equation}
where $\mat{K}$ is a set of cameras for rendering, $\mat{I}_{\vec{k}}^{input}$ and $\mat{I}_{\vec{k}}^{edit}$ are images rendered from the input mesh and the edited result, and $\mat{M}^{input}_{\vec{k}}$ is a segmentation mask of the $\mat{I}_{\vec{k}}^{input}$ which is defined as $\mat{M}_{\vec{k}}^{input}=\{(i,j)\vert \mathrm{SAM}(\mat{I}_{\vec{k}}^{input})_{ij}=1\}$. 

\subsection{Canonicalization}
\paragraph{Baselines.} For Fast-SNARF~ \cite{chen2023fast}, we use the official implementation with the default hyperparameters except for the skinning mode where we use the ``preset'' mode which uses the nearest neighbor skinning weights, instead of the original ``mlp'' mode which learns the skinning weights. This is due to the training instability with limited training data as mentioned in the main paper.

\paragraph{Ablation.}
In our ablation study, we utilize the CAPE dataset~ \cite{ma2020cape}. Since the dataset doesn't provide texture data, we employ an off-the-shelf mesh texturing tool~ \cite{richardson2023texture} to add color information to the input mesh and perform segmentation, which we find challenging to perform on the rendered geometry or normals. 

\section{Additional Qualitative Results}
In this section, we present additional qualitative results of our method. Please refer to the supplementary video for animated results.
\paragraph{Decomposing User-generated 3D Assets.}
\begin{figure}
     \centering
     \begin{subfigure}[b]{0.15\columnwidth}
         \centering
         \includegraphics[width=\textwidth]{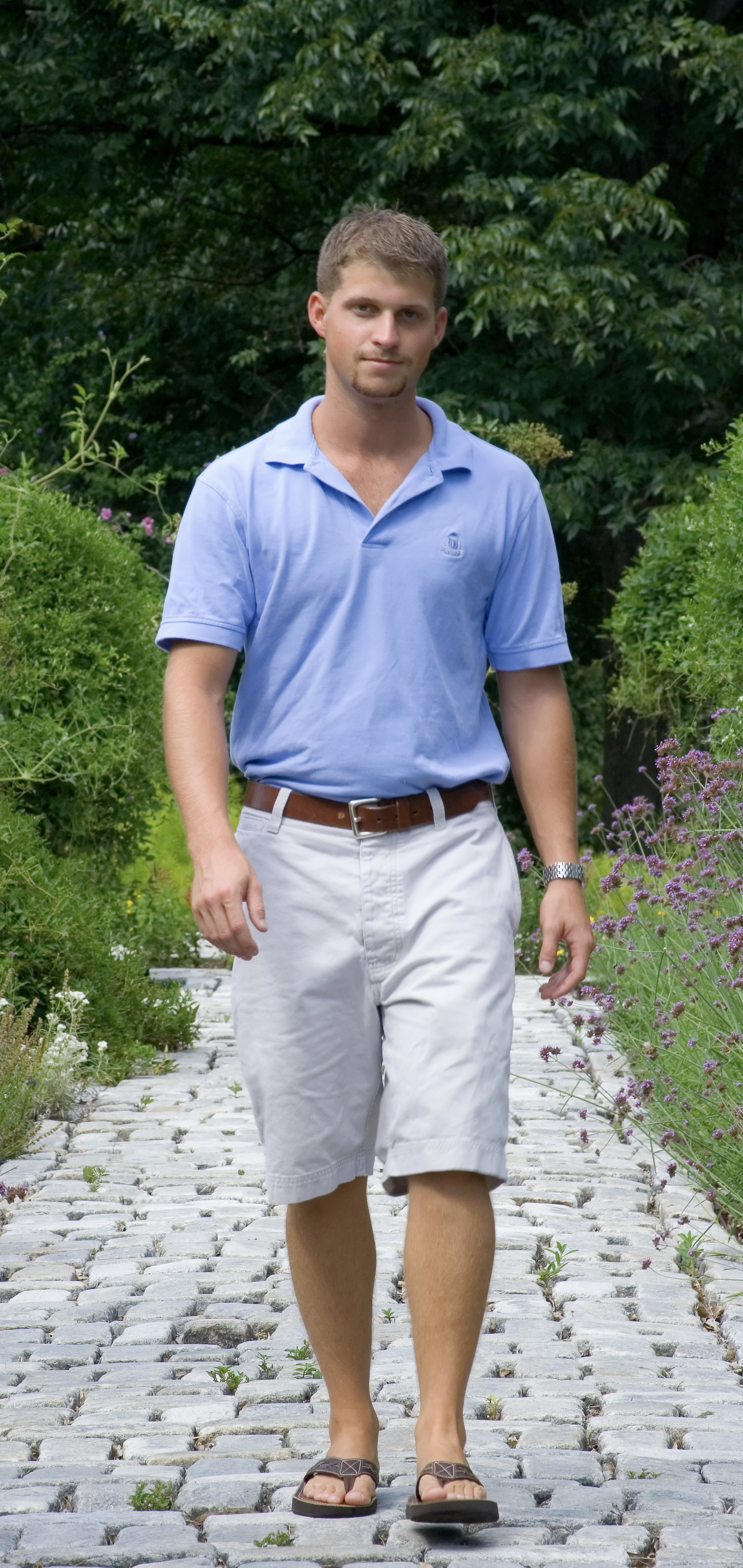}
         \caption{Image}
     \end{subfigure}
     \hspace{1em}
     \begin{subfigure}[b]{0.36\columnwidth}
         \centering
         \includegraphics[width=\textwidth]{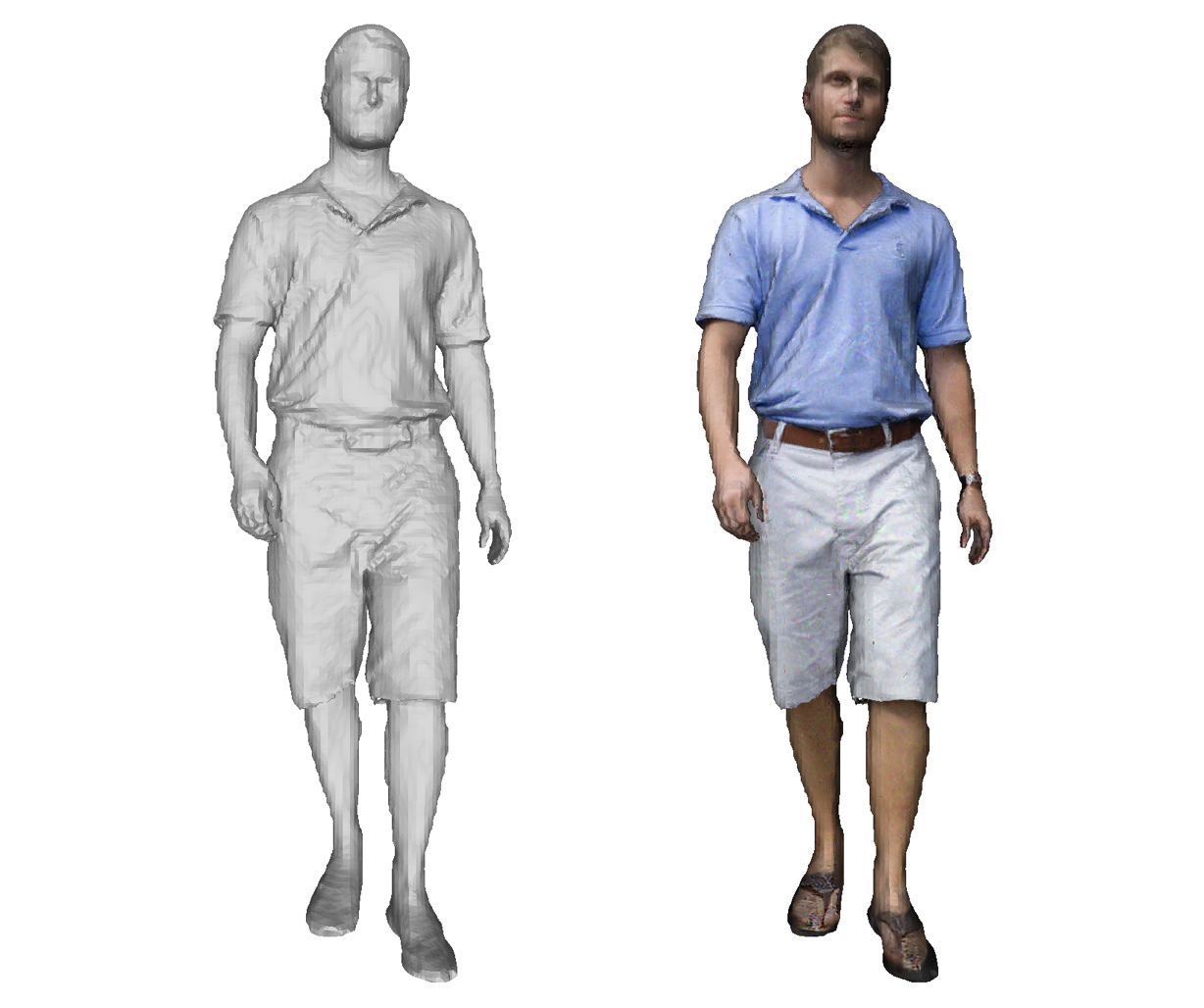}
         \caption{3D reconstruction}
     \end{subfigure}
     \begin{subfigure}[b]{0.41\columnwidth}
         \centering
         \includegraphics[width=\columnwidth]{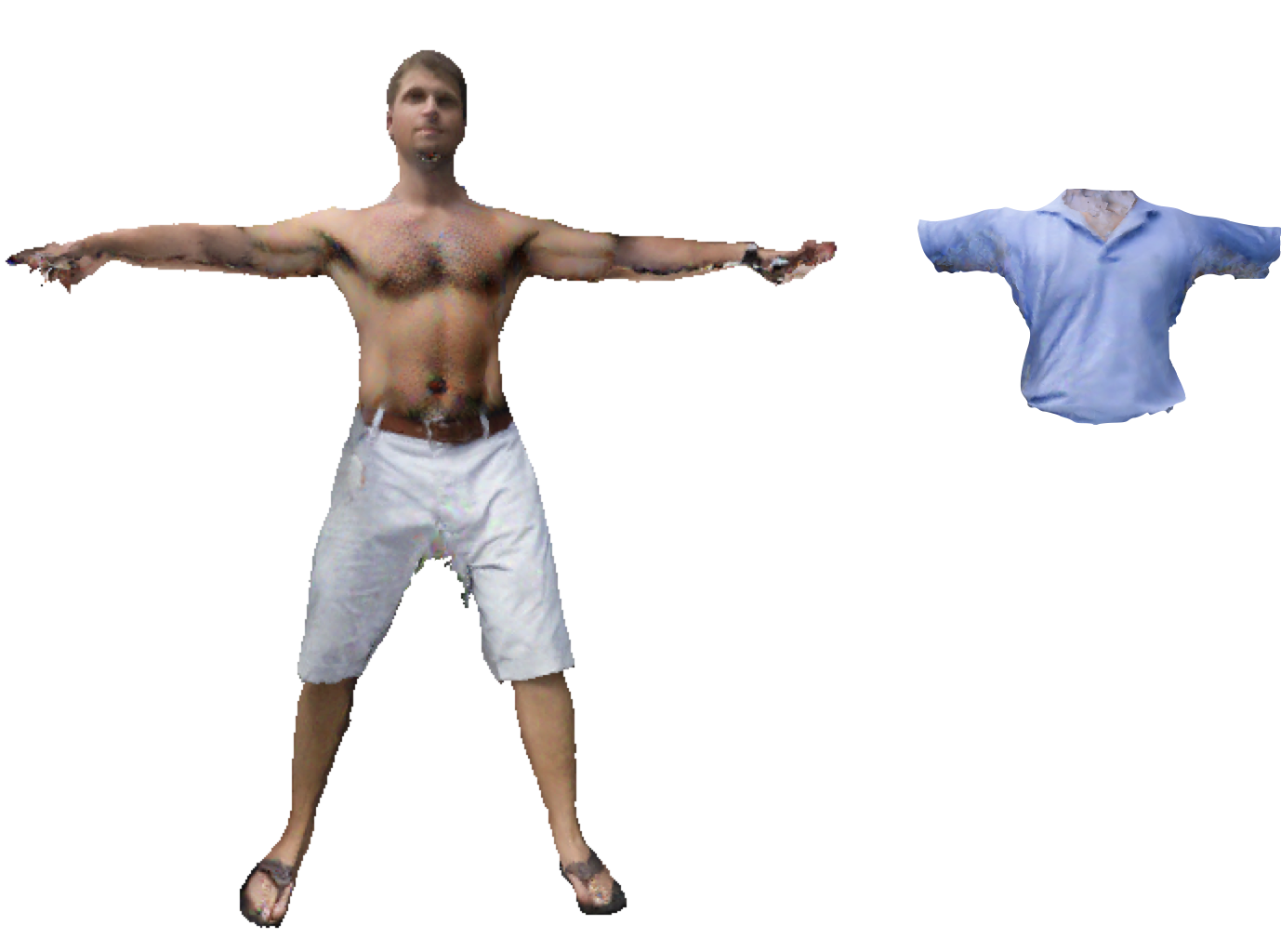}
         \caption{Decomposition}
     \end{subfigure}
        \caption{\textbf{Decomposing single-view 3D reconstructions.} Our method enables the generation of animatable layered assets from 2D images via 2D-to-3D reconstruction methods~ \cite{albahar2023humansgd}.}
\label{fig:app_recon}
\end{figure}
\begin{figure}
     \centering
     \begin{subfigure}[b]{0.38\columnwidth}
         \centering
         \includegraphics[width=\columnwidth]{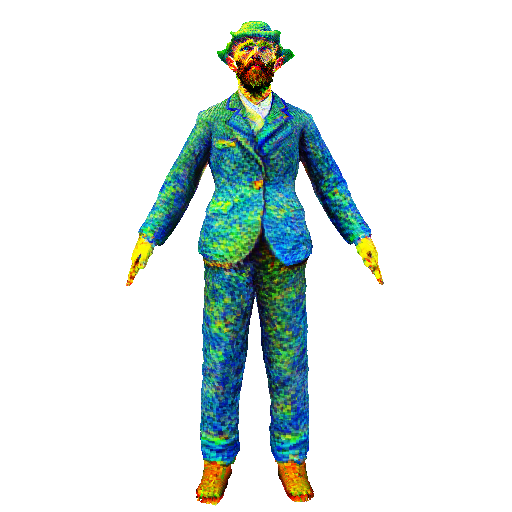}
         \caption{3D avatar}
     \end{subfigure}
     \begin{subfigure}[b]{0.57\columnwidth}
         \centering
         \includegraphics[width=\columnwidth]{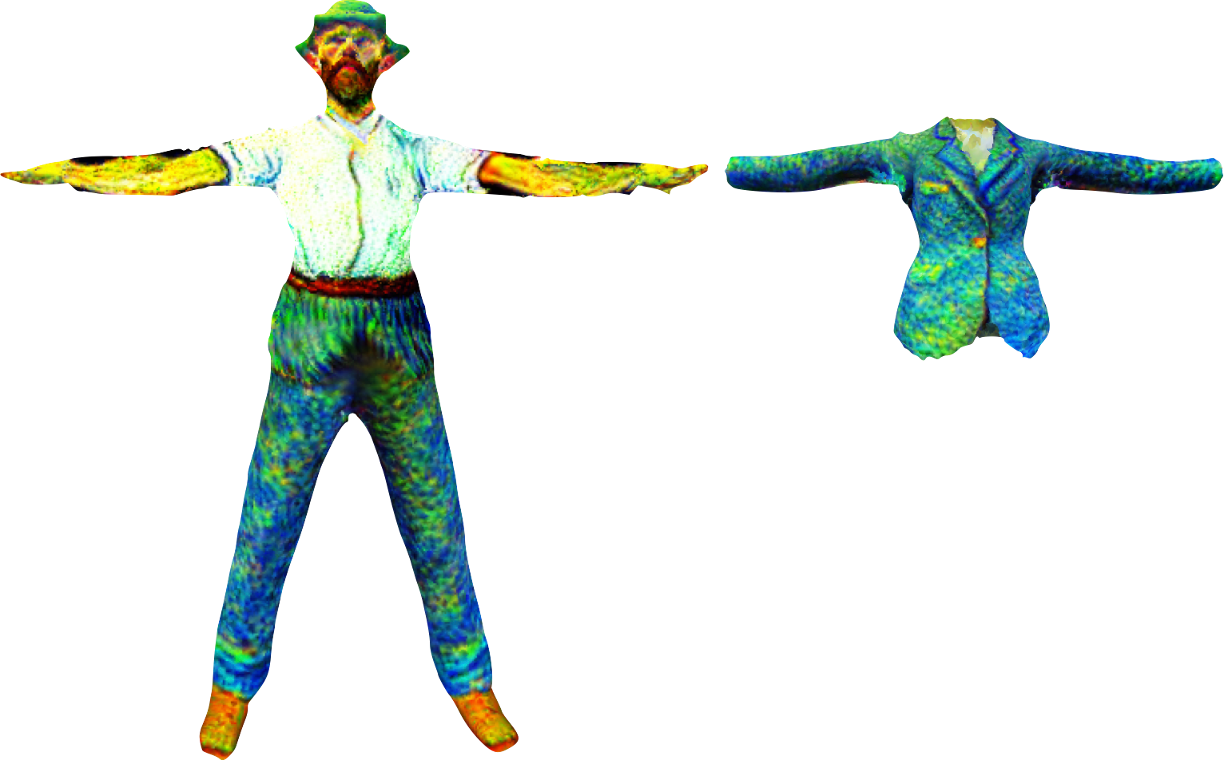}
         \caption{Decomposition}
     \end{subfigure}
        \caption{\textbf{Decomposing diffusion-generated 3D assets.} Our method enables the generation of animatable layered assets from texts via text-to-3D generation methods~ \cite{liao2024tada}. We show the decomposition result for the avatar generated with the prompts ``Vincent Van Gogh''.}
\label{fig:app_gen}
\vspace{-5px}
\end{figure}
Our method can decompose user-generated 3D assets from single-view 3D reconstruction methods~ \cite{saito2019pifu, saito2020pifuhd, xiu2022icon, xiu2023econ, alldieck2022phorhum, albahar2023humansgd, huang2024tech} or 3D avatar generation methods~ \cite{cao2023dreamavatar, kolotouros2023dreamhuman, huang2023humannorm, zhang2023avatarverse, liao2024tada}. \cref{fig:app_recon} shows the decomposition result of the 3D human mesh reconstructed from a 2D image with Human-SGD~ \cite{albahar2023humansgd} and \cref{fig:app_gen} shows the decomposition result of the 3D avatar generated from text with TADA~ \cite{liao2024tada}. These results demonstrate that GALA enables the intuitive scenario for the users to create their own reusable 3D assets from their images or text guidance.

\paragraph{Decomposition and Canonicalization.} ~\cref{fig:decomposition_supp} is an extended figure of Fig. 5 in the main paper, which shows the results of decomposition and canonicalization of input scans.

\begin{figure}[t]
\includegraphics[width=1.0\columnwidth, trim={0 0 0 0cm}]{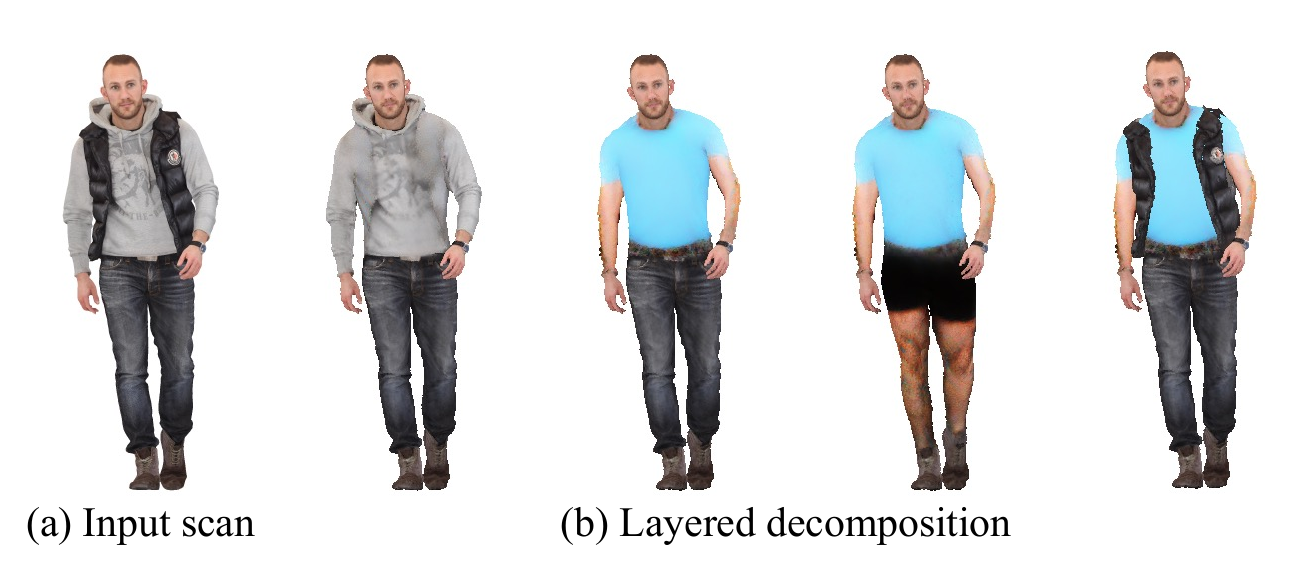}
\vspace{-5px}
\caption{\textbf{Layered decomposition.} Our method enables the layered decomposition of the input scan. Note that we can remove the specific layer of clothing by recomposing the decomposed assets.}
\label{fig:layered}
\end{figure}
\paragraph{Layered Decomposition.} ~\cref{fig:layered} is an extended figure of Fig. 1 in the main paper, which shows the strength of our method to generate ``layered'' assets by applying series of decomposition to the input scan. By composing back the decomposed assets, our method enables the decomposition of specific layers of clothing.

\begin{figure}[t]
\includegraphics[width=1.0\columnwidth, trim={0 0 0 0cm}]{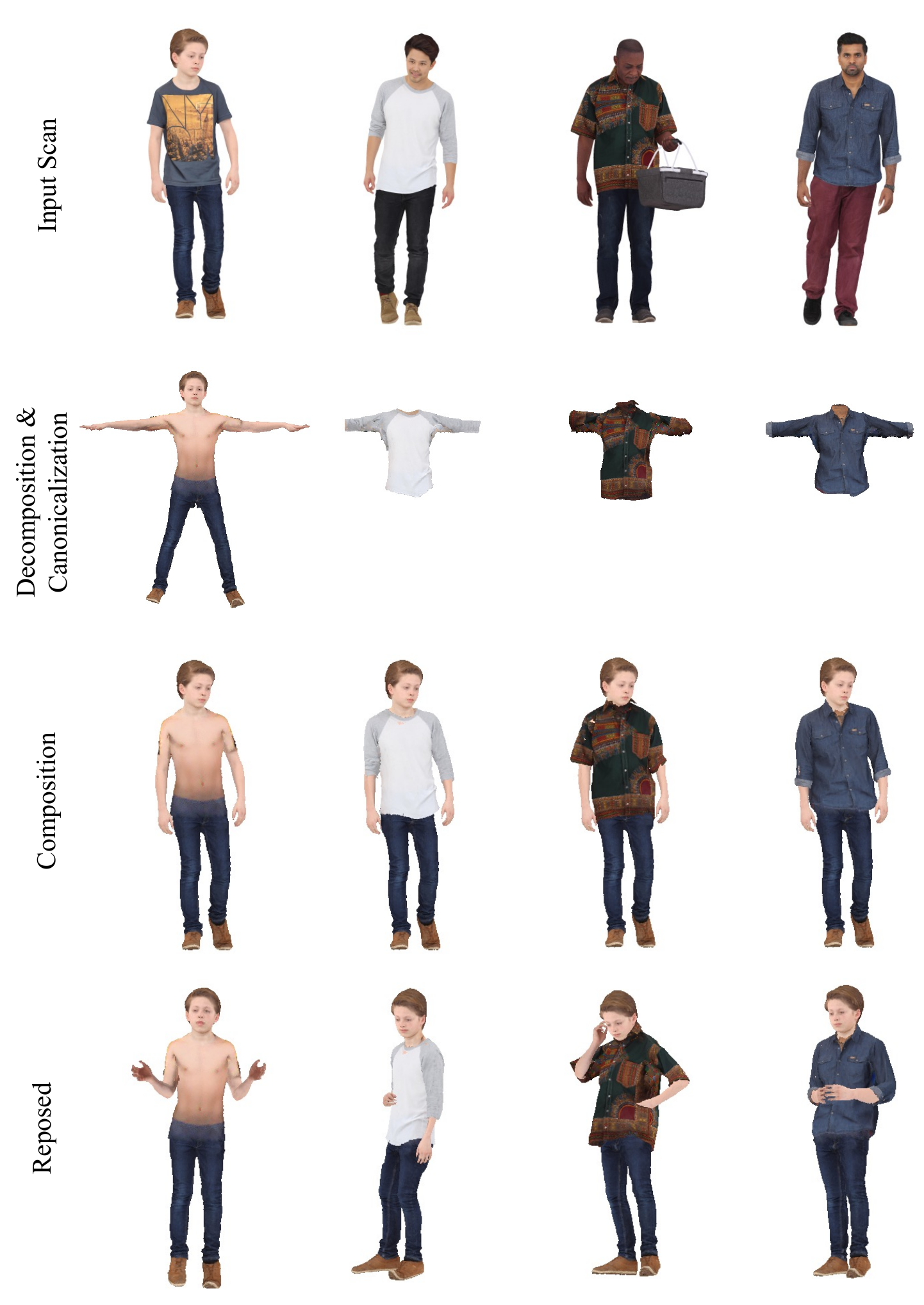}
\vspace{-5px}
\caption{\textbf{Composition.} Our method enables creation of newly-dressed avatars which are fully animatable, by combining various combinations of decomposed assets.}
\label{fig:composition}
\end{figure}
\paragraph{Composition.} ~\cref{fig:composition} is an extended figure of Fig. 1 in the main paper, depicting the ability of our method for 3D garment transfer and reposing.

\begin{figure}[t]
\includegraphics[width=1.0\columnwidth, trim={0.6cm 0 0 0cm}]{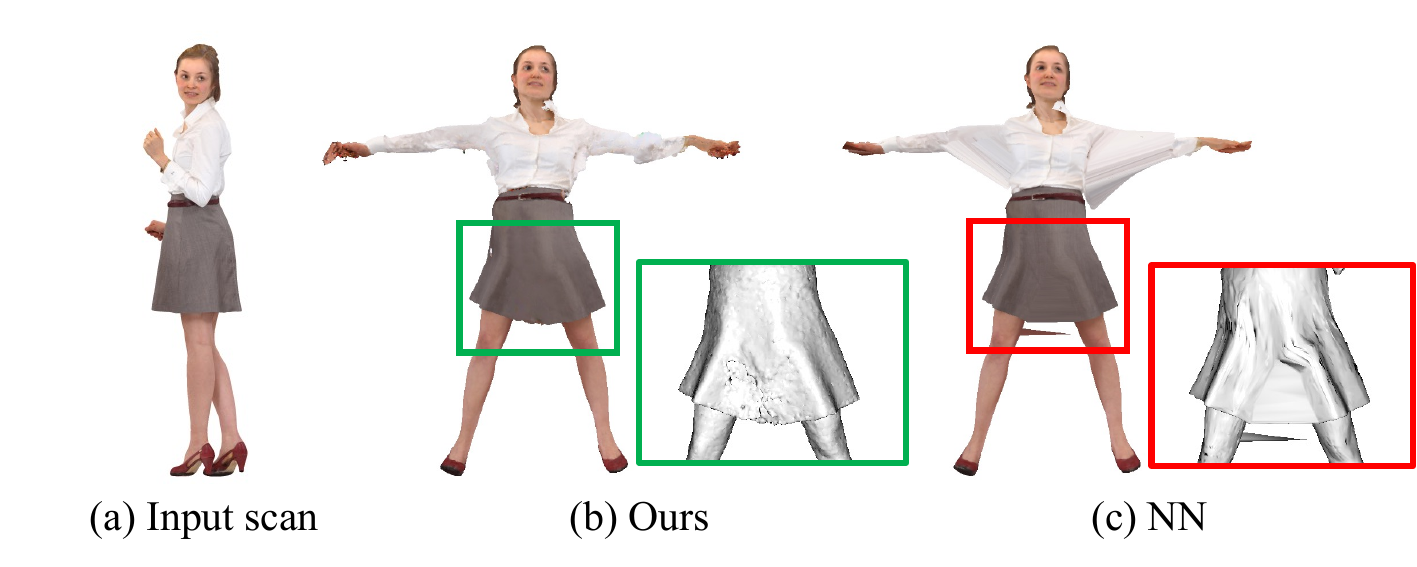}
\caption{\textbf{Loose Clothing.} Our method successfully models canonical shapes of loose clothing.}
\label{fig:loose}
\end{figure}
\paragraph{Loose Clothing.} ~\cref{fig:loose} is an extended figure of Fig. 7 in the main paper, which shows the advantage of our method for modeling canonical shapes of loose clothing compared to simple canonicalization methods~ \cite{huang2020arch, he2021arch++}. 

\begin{figure}[t]
\includegraphics[width=1.0\columnwidth, trim={0.8cm 0 0.3cm 0cm}]{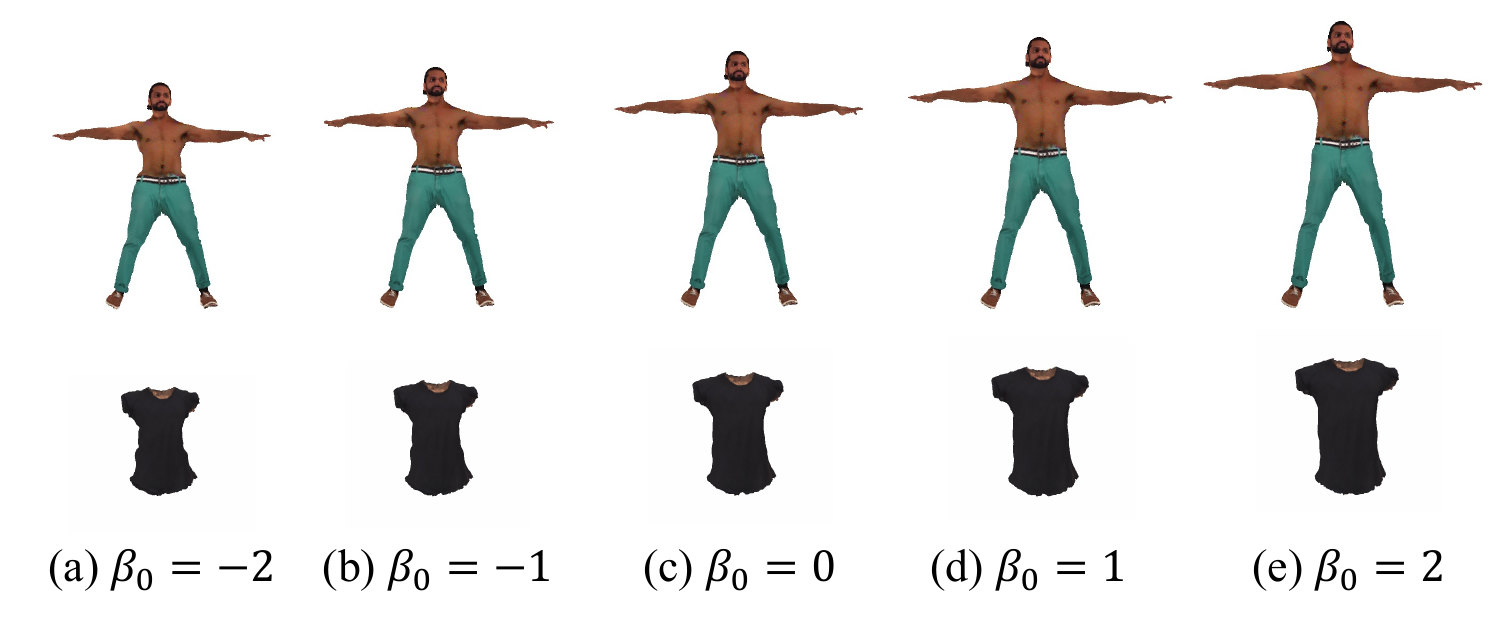}
\caption{\textbf{Size changes of decomposed assets.} Our method enables effortless size changes of decomposed assets by switching the SMPL-X shape parameters.}
\label{fig:beta}
\end{figure}
\paragraph{Size Changes.} ~\cref{fig:beta} shows the ability of our method to efficiently change the shapes of decomposed assets by altering the SMPL-X shape parameters. 

\begin{figure}[t]
\includegraphics[width=1.0\columnwidth, trim={0.5cm 0 0 0cm}]{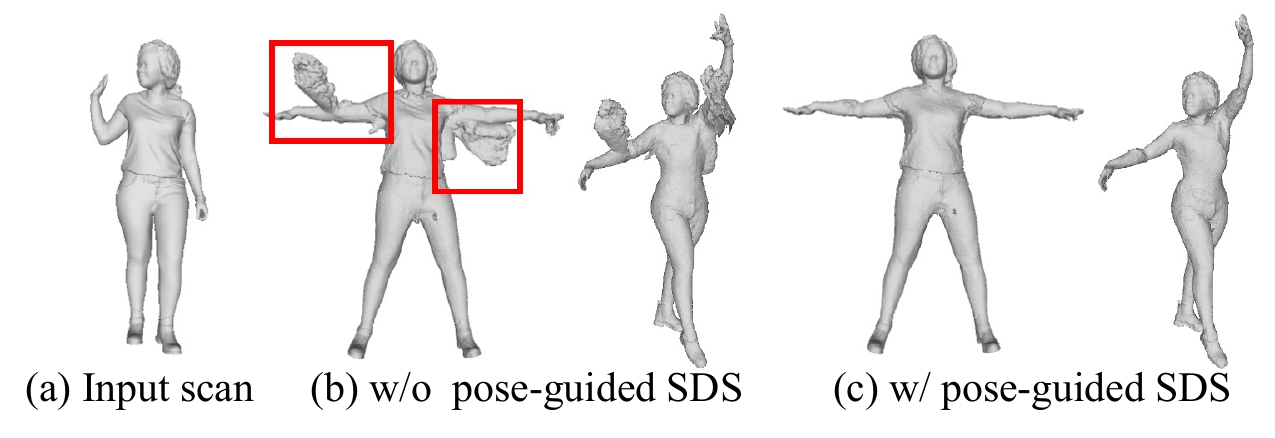}
\caption{\textbf{Canonicalization via pose-guided SDS loss.} Applying our pose-guided SDS loss in the canonical space enables robust canonicalization from a single scan.}
\label{fig:pose_guided}
\end{figure}
\paragraph{Pose-guided SDS Loss.} ~\cref{fig:pose_guided} is an extended figure of Fig. 10 in the main paper. Our pose-guided SDS loss applied in the canonical space effectively removes artifacts in the canonical shape and enables correct canonicalization from a single scan.

\section{Discussion}

\begin{table}[t]

\centering
\small{
\begin{tabular}{lcc}
\toprule
Method & IoU$\uparrow$ & Chamfer$\downarrow$ \\
\midrule
Composite & \textbf{83.59\%} & \textbf{1.184} \\
Object & 83.50 \% & 1.205 \\
\bottomrule
\end{tabular}
}
\caption{\textbf{SDS loss to composite mesh.} We show the effect of applying SDS loss to the composite mesh instead of the object mesh.}
\label{tab:sds_comp}
\end{table}

\paragraph{SDS loss to Composition Mesh.}
As mentioned in the main paper, in order to complete geometry and appearance of the object, we apply our pose-guided SDS loss to the composite mesh of human and object instead of the object mesh itself. This is due the fact that OpenPose~ \cite{cao2019openpose} ControlNet~ \cite{zhang2023controlnet} is trained to generate pose-guided human images. Hence, when given the positive prompt ``\emph{\{target object}\}'', and the negative prompt, ``a person'', it fails to exclusively generate the object without humans as shown in ~\cref{fig:sd_inference}. We also present quantitative comparison on canonicalization between applying SDS loss to the composite mesh and to the object mesh in ~\cref{tab:sds_comp}.

\begin{figure}[t]
\includegraphics[width=1.0\columnwidth, trim={0 0 0 0cm}]{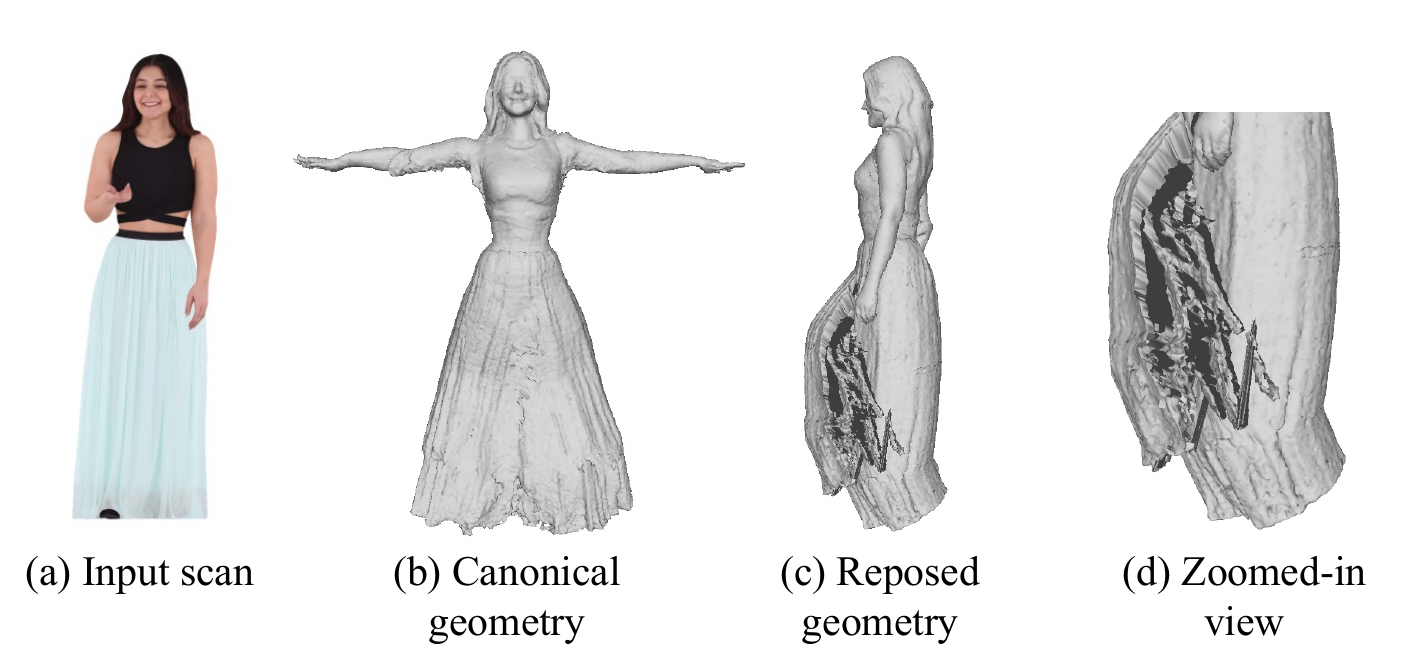}
\caption{\textbf{Failure case of reposing loose clothing.} Since our method generates static canonical shape, reposing a human with loose clothing may result in severe artifacts between the legs.}
\label{fig:limit_posedep}
\end{figure}

\begin{figure}[t]
\includegraphics[width=1.0\columnwidth, trim={0 0 0 0cm}]{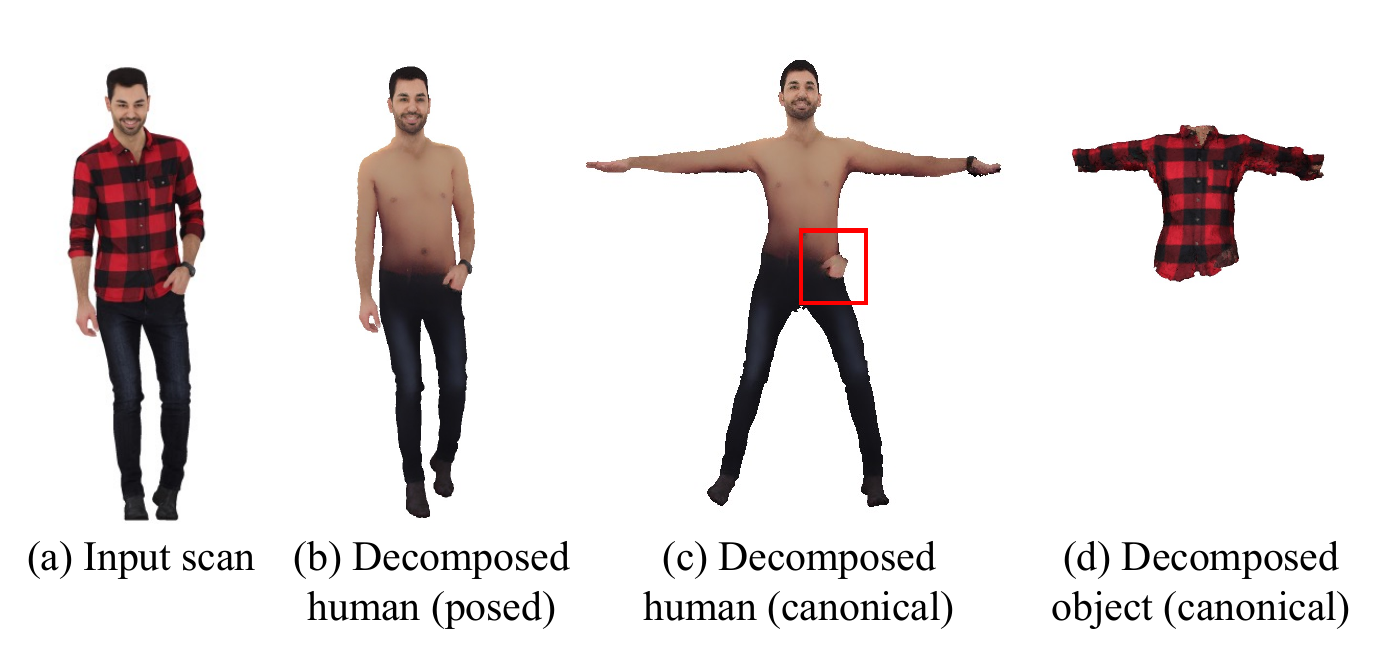}
\caption{\textbf{Failure case of canonicalization.} Our method suffers from correctly canonicalizing scans with hands in their pockets.}
\label{fig:limit_cano}
\end{figure}

\paragraph{Limitations.}
As mentioned in the main paper, GALA currently models a static canonical shape without considering pose-dependent deformations. ~\cref{fig:limit_posedep} illustrates a failure case of reposing a human with loose clothing, where severe artifacts of the dress appear between the legs. Jointly modeling pose-dependent deformation of clothing from a single scan can be a potential direction for future work. Additionally, our method may encounter challenges when canonicalizing input scans with difficult poses such as humans with theirs hands in their pockets. As shown in ~\cref{fig:limit_cano} (c), the hand partially remains inside the pocket after decomposition, limiting the reuse of the decomposed human. Nonetheless, the decomposed human can still be used in the pose of the input scan as depicted in ~\cref{fig:limit_cano} (b), and the decomposed object of ~\cref{fig:limit_cano} (d) can be utilized as any other decomposed asset.

\paragraph{Societal Impact.}
GALA decomposes a single static scan into reusable and animatable assets, \eg target apparel and the underlying human body. Similar to other recent generative models and editing methods, our method may have both positive and negative societal impacts depending on the usage. On the positive side, GALA can immediately generate diverse reusable assets from existing 3D assets that have entangled geometry, without template registration, additional scanning, or editing by 3D designers. For the metaverse applications, GALA enables users to easily digitize their assets and clothe their avatars in the virtual world. On the negative side, GALA may generate a naked underlying body for the human scan with single-layered clothing unless the input prompts are properly given. Since GALA utilizes SDS loss~ \cite{poole2023dreamfusion} to leverage the prior from the pre-trained 2D diffusion model, this problem can be alleviated via the NSFW filter. Nonetheless, there are still potential problems, \eg privacy violations, fake news, online sexual harassment, etc., like deepfake~ \cite{westerlund2019deepfake}. In our code release, we will specify the correct use of our method. We believe that the malicious use of generative models should be dealt with through both legal regulation and technology to detect misuse cases. We hope that our work invokes a serious discussion on such issues.

\begin{figure*}[t]
\centering
\includegraphics[width=1\linewidth]{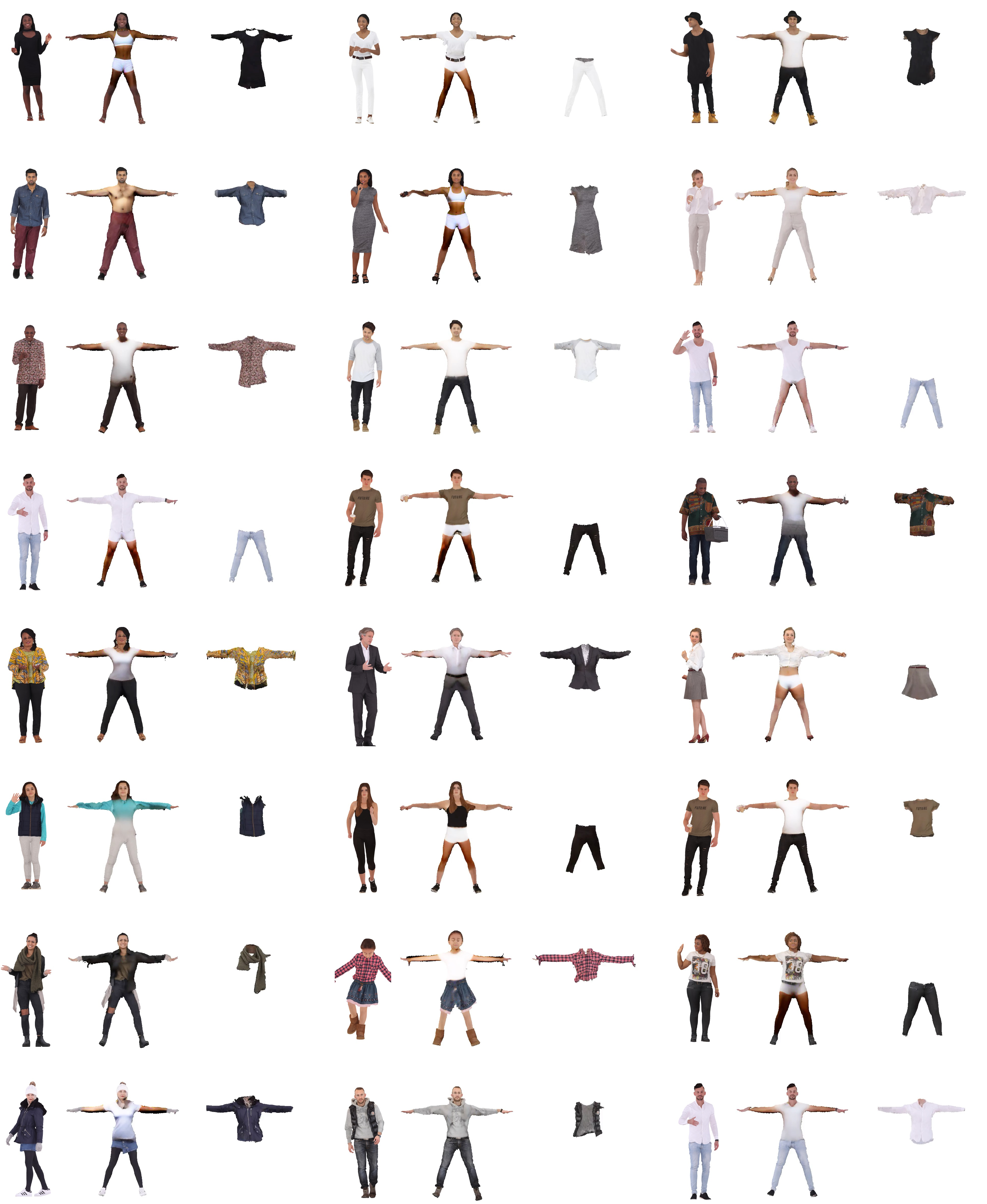}
\caption{\textbf{Decomposition and Canonicalization.} In each set, we show the decomposition and canonicalization results of the leftmost input scan.}
\label{fig:decomposition_supp}
\end{figure*}

\begin{figure*}[t]
\centering
\includegraphics[width=0.85\linewidth]{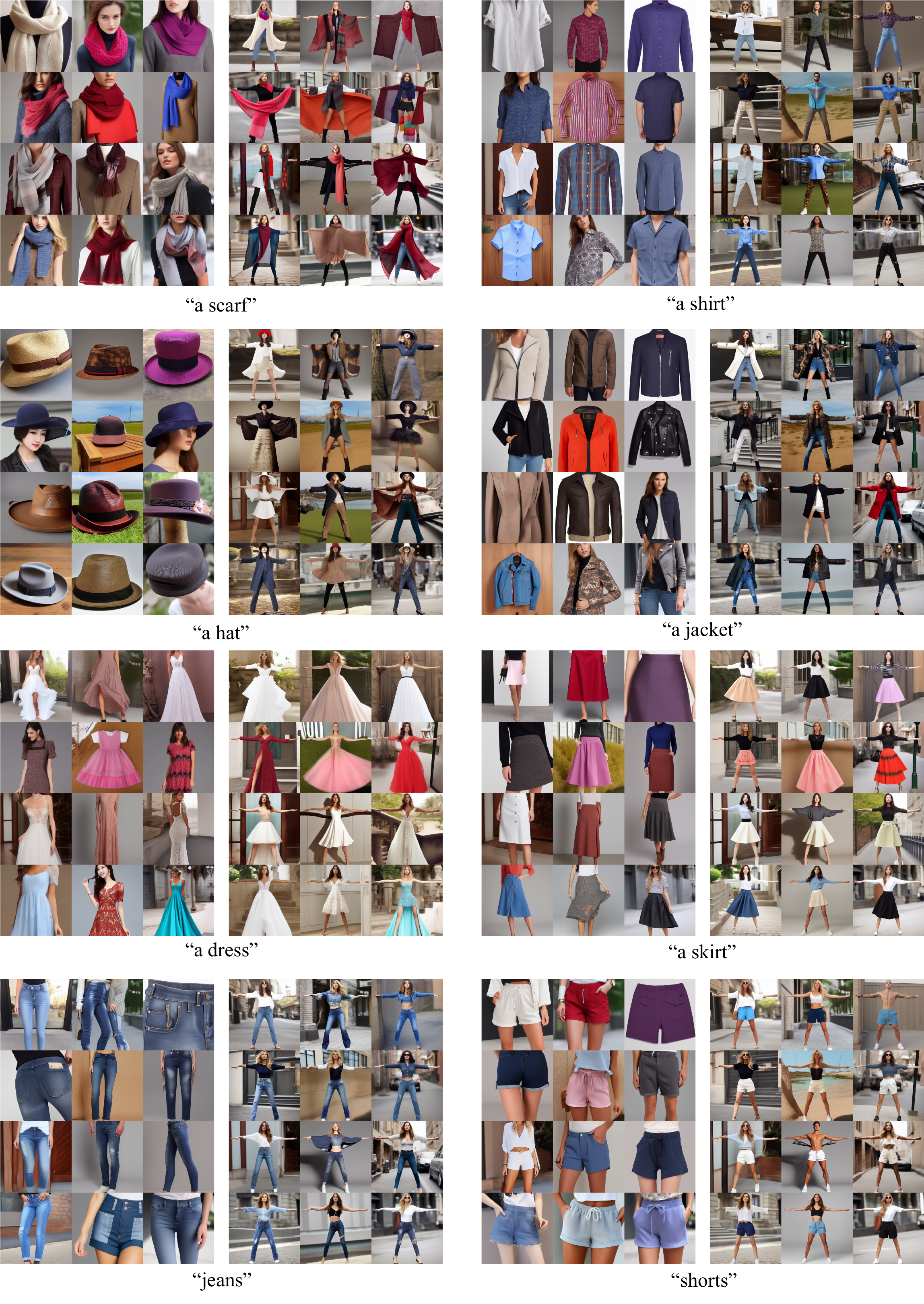}
\caption{\textbf{Pose-guided Generation.} In each set, we show the generated images of the target objects without OpenPose ControlNet on the left, and with OpenPose ControlNet on the right. Diffusion model fails to exclusively generate target objects without humans when OpenPose ControlNet is used for pose-guided SDS loss.}
\label{fig:sd_inference}
\end{figure*}

\clearpage

\fi

\end{document}